\documentclass{article}

\usepackage[preprint]{neurips_2025}

\usepackage[utf8]{inputenc} 
\usepackage[T1]{fontenc}    
\usepackage{hyperref}       
\usepackage{url}            
\usepackage{booktabs}       
\usepackage{amsfonts}       
\usepackage{nicefrac}       
\usepackage{microtype}      
\usepackage{xcolor}         

\usepackage{algorithm}
\usepackage{algpseudocode}
\usepackage[utf8]{inputenc}
\usepackage{multirow}  
\usepackage{booktabs}
\usepackage{graphicx}
\usepackage{microtype}
\usepackage{subcaption}
\usepackage{inconsolata}
\usepackage{times}
\usepackage{latexsym}
\usepackage[T1]{fontenc}
\usepackage[utf8]{inputenc}
\usepackage{microtype}
\usepackage{inconsolata}
\usepackage{graphicx}
\usepackage{colortbl}
\usepackage{amsmath}
\usepackage{booktabs}
\usepackage{amssymb}
\usepackage{adjustbox}
\usepackage{pifont}
\newcommand{\xmark}{\ding{55}}
\usepackage{bm}
\usepackage{longtable}
\usepackage{multirow}
\usepackage{multicol}
\usepackage[framemethod=TikZ]{mdframed}
\usepackage{xcolor}
\usepackage{url}
\usepackage{enumitem}
\usepackage{verbatim}
\usepackage{afterpage}
\usepackage{authblk}
\usepackage{subcaption}
\usepackage{threeparttable}
\usepackage{array}
\usepackage{enumitem}
\usepackage{graphicx}
\usepackage{booktabs}  
\usepackage{hyperref}
\usepackage{mathtools}
\usepackage{amsthm}
\usepackage{siunitx}
\usepackage{amsfonts}
\usepackage{longtable}
\usepackage{xspace} 
\usepackage{listings}
\usepackage{wrapfig}
\usepackage{graphicx}

\newcommand{\vect}[1]{\ensuremath{\bm{#1}}\xspace}

\usepackage{listings}
\title{Mobile-Bench-v2: A More Realistic and Comprehensive Benchmark for VLM-based Mobile Agents}

\author{
    \textbf{Weikai Xu$^{14*\dag}$, Zhizheng Jiang$^{24*\dag}$, Yuxuan Liu$^{34}$, Pengzhi Gao$^{4}$, Wei Liu$^{4}$, Jian Luan$^{4}$, Yuanchun Li$^{5}$, Yunxin Liu$^{5}$, Bin Wang$^{4}$, Bo An$^{1\ddagger}$} \\
  $^1$\textmd{Nanyang Technological University} 
  $^2$\textmd{Independent researchers} \\
  $^3$\textmd{Gaoling School of Artificial Intelligence, Renmin University of China} \\
  $^4$\textmd{MiLM Plus, Xiaomi Inc.}
  $^5$\textmd{Institute for AI Industry Research (AIR), Tsinghua University}\\
  xuwk266@gmail.com, gezelligheid314@gmail.com
  }

\begin{document}

\maketitle
\renewcommand{\thefootnote}{\fnsymbol{footnote}}
    \footnotetext[1]{Equal contribution.}
    \footnotetext[2]{Work done during the internship at XiaoMi.}
    \footnotetext[3]{Bo An is the corresponding author.}
\begin{abstract}
VLM-based mobile agents are increasingly popular due to their capabilities to interact with smartphone GUIs and XML-structured texts and to complete daily tasks. 
However, existing online benchmarks struggle with obtaining stable reward signals due to dynamic environmental changes. 
Offline benchmarks evaluate the agents through single-path trajectories, which stands in contrast to the inherently multi-solution characteristics of GUI tasks. 
Additionally, both types of benchmarks fail to assess whether mobile agents can handle noise or engage in proactive interactions due to a lack of noisy apps or overly full instructions during the evaluation process.
To address these limitations, we use a slot-based instruction generation method to construct a more realistic and comprehensive benchmark named Mobile-Bench-v2. 
Mobile-Bench-v2 includes a common task split, with offline multi-path evaluation to assess the agent's ability to obtain step rewards during task execution. 
It contains a noisy split based on pop-ups and ads apps, and a contaminated split named AITZ-Noise to formulate a real noisy environment. 
Furthermore, an ambiguous instruction split with preset Q\&A interactions is released to evaluate the agent's proactive interaction capabilities. 
We conduct evaluations on these splits using the single-agent framework AppAgent-v1, the multi-agent framework Mobile-Agent-v2, as well as other mobile agents such as UI-Tars and OS-Atlas. 
Code and data are available at \url{https://huggingface.co/datasets/xwk123/MobileBench-v2}.
\end{abstract}

\section{Introduction}

LLM-based mobile agents \citep{wang2023enabling,ding2024mobileagent} are increasingly popular due to their capability to interact directly with mobile Graphical User Interfaces (GUIs) and their potential to manage daily tasks autonomously. 
Unfortunately, LLM-based agents cannot well comprehend the mobile GUI structure and widget functionality, base on text such as Visual-Hierarchical, XML, HTML, or Accessible-Visited Trees.  
Recent studies \citep{ma2024coco,zhang2024mobileexperts} indicate VLMs can provide a more comprehensive understanding of GUIs.
This has led to mobile benchmark replacing foundational models with VLMs, resulting in benchmarks for end-to-end mobile tasks based on GUI pages \citep{wang2024mobileagentbench,xu2024androidlab,rawles2024androidworld}. 
\begin{table*}[ht!]
  \centering
  \small
  \begin{threeparttable}
    \caption{Comparison of Mobile-Bench-v2 to other benchmarks. 
    \textbf{Scale}: \# is the number of unique instructions on general third-party apps, average steps per instruction, and screenshots.
    \textbf{*} indicates step metrics are mis-annotated, and the tasks are not normal mobile GUI tasks.}
    \begin{tabular}{p{3.2cm} p{1.6cm} p{1cm} p{1.2cm} p{1.3cm} p{1.3cm} p{1.3cm}}
        \toprule
        \multirow{2}{*}{\textbf{Benchmarks}} & \textbf{\#} Unique\newline General Inst. & \textbf{\#} Avg\newline Steps & \textbf{\#} Screen-\newline shots & Task\newline Path & Realistic\newline Environ. & Ambigu\newline Noise \\
        \midrule
        \textsc{PixelHelp} \citep{li2020mapping} & 187 & 4.2 & $\sim$800 & Single & \xmark & \xmark \\
        \textsc{MoTIF} \citep{burns2021mobile} & 480 & 4.5 & $\sim$21K  & Single & \xmark & \xmark \\
        \textsc{AitW} \citep{rawles2024androidinthewild} & 1,539 & 6.5 & $\sim$510K  & Single & \xmark & \xmark \\
        \textsc{AitZ} \citep{zhang2024android} & 506 & 7.5 & $\sim$18K  & Single & \xmark & \xmark\\
        \textsc{AMEX} \citep{chai2024amex} & 341 & 12.8 & $\sim$104K & Single & \xmark & \xmark \\
        \textsc{ScreenSpot} \cite{cheng2024seeclick} & $\sim$1,200 & 1 & $\sim$600 & Dot & \xmark & \xmark \\
        \textsc{MobileAIBench} \citep{murthy2024mobileaibench} &\textbf{*} & \textbf{*}&\textbf{*} & Dot & \xmark & \textbf{*} \\
        \textsc{AgentBench} \citep{wang2024mobileagentbench} & 100 & 20 & $\sim$2k   & Multiple & \checkmark & \xmark \\
        \textsc{GUI Odeyssey} \citep{lu2024gui} & 7,735 & 15.4 & \textbf{*}   & Single & \xmark & \xmark \\
        \textsc{Mobile-Bench} \citep{deng2024mobile} & 832 & \textbf{*} & 14,144   & Multiple & \checkmark & \xmark \\
        \textsc{Mobile-Env} \citep{zhang2023mobile} & 150 &  \textbf{*} &  \textbf{*} & Multiple & \checkmark & \xmark \\
        \textsc{AndroidLab} \citep{xu2024androidlab} & 10.5k & 8.98 & 94.3k & Multiple & \checkmark & \xmark \\
        \textsc{AndroidWorld} \citep{rawles2024androidworld} & 116 & \textbf{*} &  \textbf{*} & Multiple & \checkmark & \xmark \\
        \midrule
        \textsc{Mobile-Bench-v2} & 12,856 & 7.28 & $\sim$48k & Both & \xmark & \checkmark \\
        \bottomrule
    \end{tabular}
    \label{benchmarks}
  \end{threeparttable}
  \vspace{-0.6cm}
\end{table*}

Existing VLM-based Agent benchmarks can be divided into two categories:
(1) \textbf{Online evaluation} involves the agent executing operations on a real device based on the user's high-level instructions. These benchmarks directly determine the success rate by checking the widget values in the final GUI and allow agents to complete tasks through multi-path. However, due to the instability of the device environment, such as OS updates, APP updates, and user preference records, online benchmarks \citep{murthy2024mobileaibench,deng2024mobile,wang2024mobileagentbench} step rewards are fluctuating and unstable. 
Meanwhile, a considerable portion of task evaluation relies on the process rather than solely on the outcome. 
(2) \textbf{Offline evaluation} uses static datasets where the golden path is pre-executed on the device, with actions and screenshots saved offline. The agent generates the current action based on each step's GUI, instructions and action history, which action trajectory is formulated as single-path. Although offline benchmarks \citep{chai2024amex,cheng2024seeclick,rawles2024androidinthewild} are more convenient, considering the diversity of agent task solutions, the agent's good performance may only represent a good fit to the preferences encoded in the current benchmark annotations, but does not necessarily indicate robustness or the ability to handle multi-path solutions. 
More critically, benchmarks such as MobileAgentBench \citep{wang2024mobileagentbench} and AutoDroid \citep{wen2024autodroid} are constructed on real devices and evaluated within Google apps using the Android Accessibility Service; these apps feature overly clean pages without task-irrelevant ads, buttons, and pop-ups.
In real-world scenarios, users may not be able to provide such precise and full instructions all at once \citep{wang2024learning}. 
Overall, existing benchmarks have following limitations, including a lack of multi-path evaluation, overly clean testing environments, and overly explicit instructions.

To address the above problems, we form a new benchmark named Mobile-Bench-v2 based on Mobile-Bench \citep{deng2024mobile}. Mobile-Bench-v2 has the following improvements:
(1) \textbf{Instruction Generation Method}: Based on the random walk graph-structured corpus of Mobile3M \citep{wu2024mobilevlm}, we use the GIAS (Generating Instructions From GUI Action Sequences) to construct 12k instructions. 
Each instruction is generated via task templates and slot information based on the corresponding trajectories. 
Slot information is derived from keynode actions in the execution trajectory, which can be used as stable step reward signals during the evaluation stage. 
(2) \textbf{Offline Multi-path Evaluation}: Combining the advantages of both online and offline environments, we propose a multi-path evaluation approach. 
We allow the agent to execute in a single-path manner and compare the result with the golden path. Alternatively, the agent is also allowed to perform action search within the graph corpus and accumulate step rewards, as it does during online evaluation. 
Both approaches are jointly adopted to comprehensively evaluate the agent's performance.
(3) \textbf{Realistic Noisy Environment}: We collect an additional sub-dataset named Mobile-Bench-Noisy to build realistic noisy evaluation. Several apps with substantial ads and pop-ups are specifically selected to collect this type of trajectory. Additionally, we contaminate AITZ by inserting ads into original normal trajectories to build AITZ-Noise. 
(4) \textbf{Active Interactive Evaluation}: We construct a sub-dataset named Mobile-Bench-Ambiguous, which allows agents to ask when necessary during task execution. Full instructions are pre-constructed and then progressively simplified into ambiguous instructions through slot-based extraction.  Questions and answers are built based on slot information and then assigned to the corresponding GUI.
Since none of the existing agent frameworks incorporates a dedicated questioning stage, we allow the agent to decide whether to raise a question before making a decision. 
Questions related to action decisions or UI element functionalities are rejected to ensure the evaluation's fairness.




Overall, our work makes four contributions:

$\bullet$ We construct a multi-path offline benchmark named Mobile-Bench-V2 based on Mobile3M's graph structure corpus and propose a slot-based instruction generation method named GIAS. 

$\bullet$ We propose an offline multi-path evaluation method and leverage slot-based key node annotations to enable stable assessment of step rewards.

$\bullet$ We introduce Mobile-Bench-Noisy to support realistic noisy evaluation by collecting data from noisy apps, enabling robust assessment under challenging environments. 

$\bullet$ We propose Mobile-Bench-Ambiguous to facilitate active interactive evaluation, where agents are allowed to ask clarification questions during execution.

\vspace{-0.2cm}
\begin{figure*}[ht]
  \centering
  \includegraphics[width=0.95\textwidth]{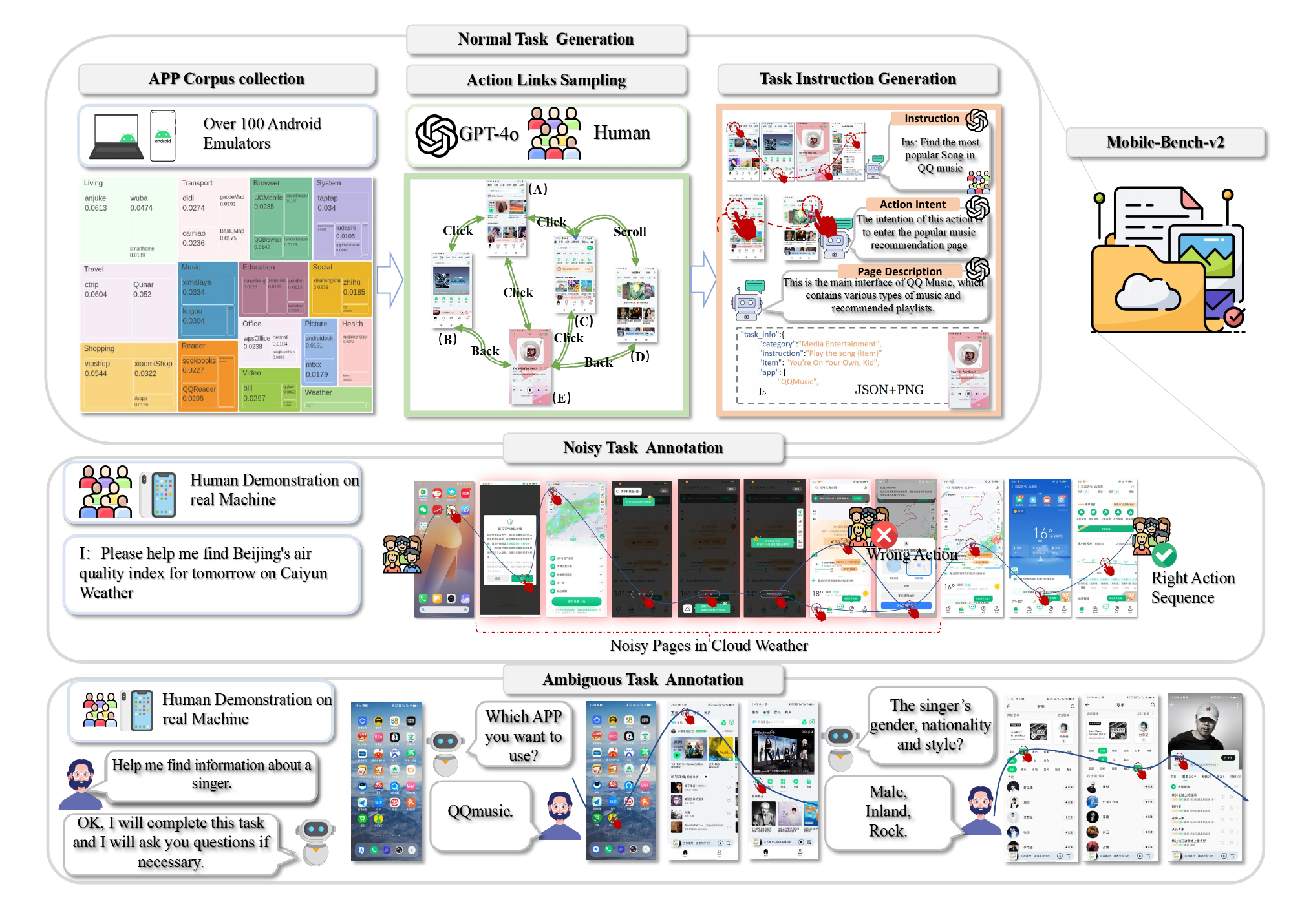}
  \caption{The \textbf{Mobile-Bench-v2} includes three types of tasks: \textbf{Common-split, Noisy-split, and Ambiguous-split}, and demonstrates the process of instruction generation and manual annotation for each task. In Noisy-split, the GUIs with red shading represent noise. }
  \label{fig:mobile_main}
\end{figure*}
\vspace{-0.6cm}
\section{Related work}
\subsection{Mobile Agents}
Large language models \citep{achiam2023gpt} emerge as autonomous agents \citep{lieffects,wen2023empowering} in the mobile domain and garner considerable attention. 
With the rapid development of vision-language models (VLMs), multimodal researchers build mobile GUI agents \citep{yang2023appagent,zheng2024gpt} and multi-agent frameworks \citep{ding2024mobileagent,li2024appagent,wang2024mobile} based on closed-source VLMs. 
Meanwhile, some researchers focus on training agents with stronger element grounding \citep{cheng2024seeclick,hong2024cogagent,wu2024atlas}, page navigation \citep{niu2024screenagent,lu2024gui,gou2024navigating}, GUI understanding \citep{chai2024amex,you2024ferret,baechler2024screenai} and task planning capabilities \citep{zhang2024llamatouch,nong2024mobileflow,xu2024aguvis} based on open-source VLMs. 
In addition, Digirl \citep{bai2024digirl} and 
Distrl \citep{wang2024distrl} uses joint online and offline reinforcement learning to enhance the generalization of mobile agents and mitigate performance degradation when facing APP updates and unseen APPs. 
Some researchers \citep{qinghong2024showui} explore optimizing VLM structures. 
For example, Dorka \citep{dorka2024training} optimizes the encoder by incorporating historical images and actions as input.
\subsection{Mobile Agent Benchmarks}
As shown in Table \ref{benchmarks}, AndroidEnv \citep{toyama2021androidenv} and MobilEnv \citep{zhang2023mobile} are the first to create LLM agent evaluation environments based on reinforcement learning. 
Mobile-Bench \citep{deng2024mobile} and AppBench \citep{wang2024appbench} introduce online benchmarks combining API and GUI, while MobileAgentBench \citep{wang2024mobileagentbench} establishes the first fully automated multimodal benchmark for VLM-based GUI agents. More offline benchmarks \citep{li2020mapping,burns2021mobile,murthy2024mobileaibench} are released, which are primarily categorized into GUI understanding and task-oriented.
(1) For task-oriented benchmarks, AITW \citep{rawles2024androidinthewild} and AITZ \citep{zhang2024android} create large-scale benchmarks based on Google apps, while AMEX \citep{chai2024amex} supplements these benchmarks by adding data for GUI understanding with similar app types.
ScreenSpot \citep{cheng2024seeclick}, Mobile3M \citep{wu2024mobilevlm}, and GUIOdyssey \citep{lu2024gui} focus on more granular element grounding and task planning. 
(2) Rico \citep{deka2017rico} is the first non-annotated GUI corpus, followed by ScreenQA \citep{hsiao2022screenqa}, Widget Caption \citep{li2020widget}, and Screen2words \citep{wang2021screen2words}, which is for Q\&A, widget understanding, and page summarization. 
Subsequently, Mind2web \citep{deng2024mind2web} incorporates additional GUI data of varying sizes, and Meta-GUI \citep{sun2022meta} provides tasks for multi-round dialogues.

\section{Mobile-Bench-v2}
\subsection{Mobile Task Formulation} 
For mobile agents, there are four essential capabilities: (i) \textbf{Overall Planning} to determine the action step sequences. (ii) \textbf{Action Thought} to produce an action description at each step (\textit{e.g.,} ``open the flight detail page''),  (iii) \textbf{Element Grounding} to identify a widget (\textit{e.g.,} ``[Click]($x_1$,$y_1$)'') on the GUI, (iiii) \textbf{Action Reflection} to determine whether the next GUI matches the expectation.
Given a mobile screenshot $\mathcal{S}$ (\textit{e.g.,} a Ctrip screenshot on Android) and a task $\mathcal{T}$ (\textit{e.g.,} \emph{``Book a flight ticket from Chengdu to Beijing Sep.15 for me.''}), a GUI agent should generate a sequence of executable actions. Specifically, at time step $t$, the agent should select an action $\vect{a}_t$ from the action space $\mathcal{A}$, which includes three types of actions: (1) Click. (2) Scroll. (3) Input.
\begin{equation}
\small
\hat{a}_t =
 \begin{cases}
    [x_1,y_1,x_2,y_2], & \hat{a}_t \in \mathrm{Click} \\
    \left\{ \uparrow, \downarrow, \leftarrow, \rightarrow \right\}, & \hat{a}_t \in \mathrm{Scroll} \\
    text, & \hat{a}_t \in \mathrm{Type} 
\end{cases}
\end{equation}
Based on the current environment observation $\mathcal{S}_t$, the action history $\mathcal{H}_{1:t-1}$=$\{\hat{a}_1, \hat{a}_2, ..., \hat{a}_{t-1}\}$, and the last step refection $\vect{f}_{t-1}$, the GUI agent will generate plan $\mathcal{P}_t$: 
\begin{equation}
\small
    \mathcal{P}_t = \Big\{\hat{a}^{(1)}_{t} \cdots \hat{a}^{(n)}_{t}  \mid (\hat{a}_1, \cdots, \hat{s}_{t-1}), f_{t-1}, \mathcal{S}_t \Big\}
\end{equation}
where $\mathcal{P}_t$ represents the planning of the next $n$ actions starting from the current step. The environment observation $\mathcal{S}_t$ comprises an HTML document ${text}_t$ and a mobile screenshot ${image}_t$. If the current step is the first step of the entire task, the overall plan $\mathcal{P}_\mathcal{T}$ can be expressed as:
\begin{equation}
\small
    \mathcal{P}_\mathcal{T} = \Big\{\mathcal{A}_{1:n} \mid \vect{C}_a, \arg\max_{\{\mathcal{T}_{i_j}\}^k} \sum_{j=1}^{k} \textsc{sim}(\mathcal{T}_i, \mathcal{T}_{i_j})\Big\}
\end{equation}
where $\vect{C}_a$ is the corpus of each APP, which is collected by the GUI agent's random walk. $\sum_{j=1}^{k} \textsc{sim}(\mathcal{T}_i, \mathcal{T}_{i_j})$ is the $top_k$ recall of the similarity comparison between the current task $\mathcal{T}_i$ and backup task $\mathcal{T}_{i_j}$.

\subsection{Data Construction} 

\noindent \textbf{Generating common instructions from action trajectories.}
When constructing the Mobile3M graph corpus, the key challenge is annotating instructions for each trajectory that closely align with the intended actions. 
Building on Netscape \citep{murty2024nnetscape}'s fine-tuning of web agents to eliminate redundant actions from action sequences, two key points for pairing trajectories and instructions are the \textbf{Intent Understanding} and the \textbf{Slot Matching} between different GUIs: 
(i) Using intents behind actions instead of themselves can reduce ambiguity because coordinate-based actions without GUI pages cannot accurately reproduce the action scenes.  
On the one hand, buttons with the same textview may have entirely different meanings on the same GUI. 
In Figure \ref{same_button}, the same \textit{`plus'} button represents adding \textit{`Hazelnut Latte'} and \textit{`Cookie Mocha'} respectively. 
On the other hand, UI elements without textual descriptions fail to reflect their underlying functionality.
For example, the sorting button in Ctrip is a triangle-shaped graphical button, recording the shape of this button is much less clear than capturing its sorting functionality. 
(ii) Filling predefined instruction templates with slot information can serve as a reward signal at key nodes. 
GUI agent tasks are inherently multi-path in nature. 
Single-path annotations with unstable preference may lead to a sharp performance drop on unseen tasks and further compromise the accuracy of performance evaluation.
Slot filling allows a single instruction template to match multiple distinct trajectories, as long as they share the same key nodes. 
This also serves as the foundation for multi-path evaluation.
The details of Mobile3m are described in Appendix \ref{appendix:randomwalk}, and data distribution is shown in Figure \ref{mobile3m_count}.

Building on the above findings, we propose a slot-based instruction annotation method named \textbf{GIAS} (Generating Instructions From Mobile UI Action Sequences), which is shown in Figure \ref{fig:mobile_main}. The whole process is as follows: (1) multi-path sampling based on fixed start and end GUIs; (2) GUI content annotation; (3) action intent inference; (4) extracting slot information from GUI changes; (5) filling instruction templates with the slot; (6) deduplication and simplification. The entire process is explained in detail in Algorithm \ref{alg:GIAS}. Specifically, we choose paths that start from nodes with the same name in Mobile3M and end at homogeneous nodes with different names (Homogeneous nodes refer to pages whose similarity or the number of identical UI elements exceeds the threshold \citep{lu2006field}).  
Considering its diversity, each trajectory includes at least two different types of actions and minimises the proportion of intermediate homogeneous pages. 
Throughout the entire annotation process, only the verification step involves a closed-source model, while all other steps are performed by open-source models without any human intervention.
More details on GIAS are in Appendix \ref{appendix:data}. 
\begin{algorithm}
\small
\caption{GIAS Algorithm}
\label{alg:GIAS}
\begin{algorithmic}[1]
\Require
    Start Page, $P_0$;
    End Page, $P_t$;
    Trajectory $\sigma$;
    Page Description $\mathcal{D}$;
    GUI Pages, $s$;
    Action, $a$;
    Action Intent, $T$;
    Page Slot, $C$;
    Instruction, $I$;
    Task, $\mathcal{T}$;   
\Ensure
    Prompt, $P$;
    Few Shot Cases, $F_S$;
    Verified Flag, $\vect{Q}$;
    Instruction Templates, $\gamma$; 
\State Select \[\sigma_i = \{s_{i_1}, s_{i_2}, \dots, s_{i_{t-1}}\}, \quad s_{i_j} \sim P_{0:t}, \quad \text{for } j = 1, \dots, t-1
\]

\For {each $s_{ij} \in \sigma_i$}
    \For { $j = 0:t$ }
        \State $\mathcal{D}_{s_{ij}} \leftarrow \mathrm{VLM}(s_{ij}, P_{t})$  \Comment{Get text descriptions from GUIs}
        \State $T_{i_{j:j+1}} \leftarrow \mathcal{D}_{s_{ij}}, \mathcal{D}_{s_{i, j+1}}, a(s_{ij} \to s_{i, j+1})$ \Comment{Get indent between two actions}
        \State $C_{i_{j:j+1}} \leftarrow \mathcal{D}_{s_{ij}}, \mathcal{D}_{s_{i, j+1}}$  \Comment{Get Page Slot based on pre-settings}
        \State $I_{ij} \leftarrow (C_{i:t}, T_{i:t}) \sim \text{Uniform}(\mathcal{\gamma})$ \Comment{Fill Templates with the slot} 
    \EndFor
\EndFor
\For{each pair of instructions $(I_i, I_j)$ in $\mathcal{I}$}
    \If{$\mathrm{Sim}(I_i, I_j) \geq \tau$}
        \State Discard $I_j$ from $\mathcal{I}$    \Comment{Make sure there are no highly similar instructions} 
    \EndIf
\EndFor

\State $\mathcal{T}, Q \leftarrow \mathrm{Veri} \ \{\mathcal{I}, P, F_S\}$ 
\Comment{for all Trajectories $\sigma_i$, ensuring no redundant steps}

\State \Return $\mathcal{T}$
\end{algorithmic} 
\end{algorithm}
\vspace{-0.3cm}

\begin{wrapfigure}{r}{0.4\columnwidth}
  \includegraphics[width=0.38\columnwidth]{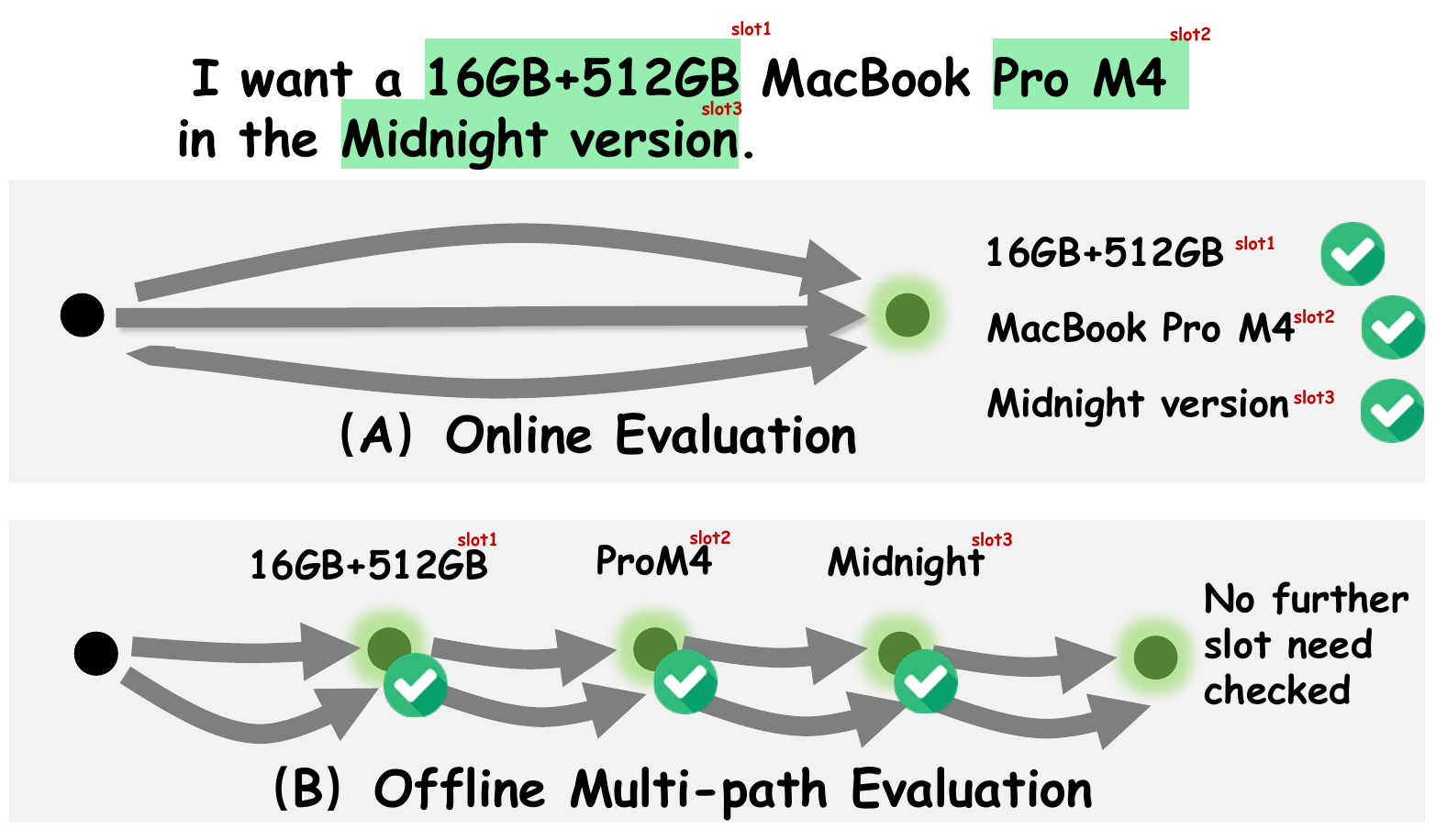}
  \caption{Unlike Online Evaluation, offline multi-path evaluation checks both process and final GUI as reward signals.}
  \label{fig:offine_multipath}
\end{wrapfigure}

\noindent \textbf{Noisy app and ambiguous instruction data.}
Mobile-Bench-Noisy is primarily derived from manual annotation and contamination in existing data: (1) For manual annotation, we select apps from third-party markets; these apps contain unavoidable ads and pop-ups. 
When performing actions on these apps, we do not handle the following scenarios in advance: login, update, permission settings, ad pop-ups, and VIP subscriptions. In some instructions, to test the agent's response in unexpected situations, we deliberately click on ad pages incorrectly to see if it can recover from divergent paths. 
All noisy steps and their redirect pages are additionally marked, making this subset adaptable for agents with a rollback mechanism as well. 
(2) For data contamination, we randomly insert at least one ad (randomly collected from the Google Store app) into AITZ \citep{zhang2024android} trajectories, which is a high-quality subset of AITW \citep{rawles2024androidinthewild}.   
For Mobile-Bench-Ambiguous, we first construct the full instructions, annotate action trajectories, and remove slot information to build ambiguous instructions. Multiple sets of interactive Q\&A are assigned to corresponding GUIs. For example, as shown in figure \ref{fig:offine_multipath}, the full instruction is: `\textit{I want a 16GB + 512GB MacBook Pro M4 in the Midnight version.}', while the ambiguous instruction is: `\textit{I want to buy a MacBook.}'. Information such as ``16GB + 512GB'', ``Pro M4'', and ``Midnight version'' are treated as three slots, each assigned to the corresponding GUI for step reward.
More details can be seen in the appendix \ref{appendix:Ambiguous Data}. 
\begin{figure*}[ht]
  \centering
  \includegraphics[width=\textwidth]{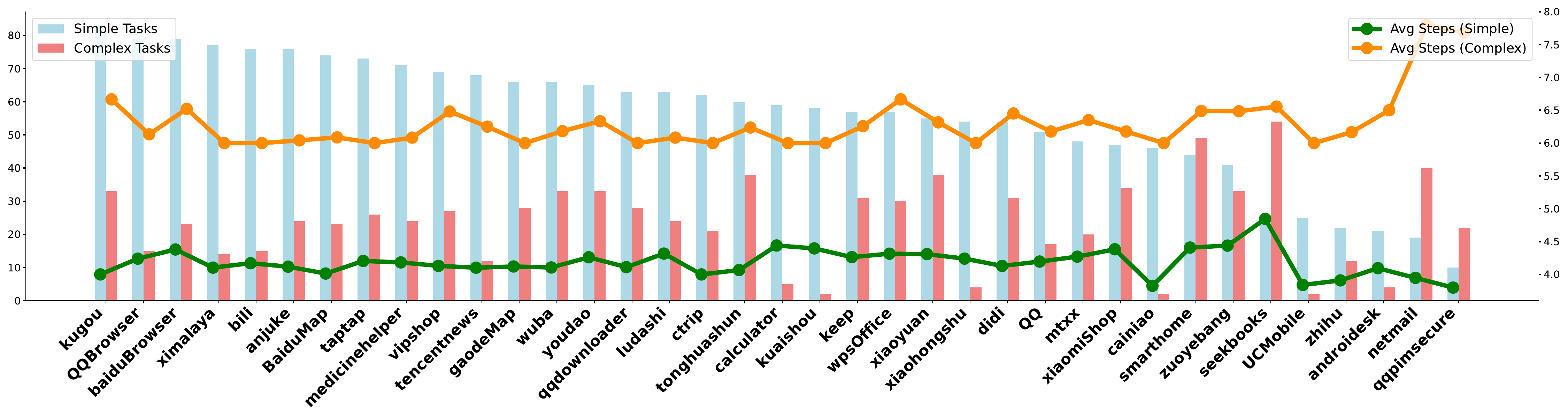}
  \caption{The \textbf{task distribution} chart is sorted by the number of simple tasks in descending order. The average steps for both simple and complex tasks in each app remain relatively balanced.}
  \label{fig:task_distribution}
\end{figure*}
\subsection{Data Statistics} 
The apps and categories in Mobile-Bench-v2 remain consistent with Mobile3M, comprising 15 categories and 49 apps, with each category containing at least the top three apps by download volume. 
As shown on the left part of Figure \ref{fig:app_fan}, QQMusic and Kugou account for over 70\% of the monthly downloads in the music app category, which are selected as representative music apps in our dataset.
Common split includes 12,854 instructions and 800 templates generated by GIAS, which are divided by action steps: simple tasks (1-6 steps) and complex tasks (7-15 steps). 
As shown in the middle part of Figure \ref{fig:app_fan}, there are 9,620 simple tasks with an average of 5.62 steps and 3,234 complex tasks with an average of 8.21 steps. 
Figure \ref{fig:task_distribution} shows the task distribution. 
We strive for a difficult balance, but some apps, like \textit{Netmail} (an email tool), still imbalance because sending an email often entails multiple steps to complete essential information fields, resulting in a relatively long interaction process that cannot be easily classified as a simple task.
From a categorical perspective, shopping apps (\textit{DuApp}) have a higher proportion of complex tasks compared to simple tasks. 
In contrast, since \textit{Baicizhan} features a clean page and straightforward functionality (\textit{vocabulary learning}), constructing task instructions from templates with fewer slots.
The noisy and ambiguous splits each contain 100 instructions, while these noisy data come from another 20 highly noisy apps and each trajectory in the ambiguous split includes at least 5 additional manually constructed Q\&A. 
As shown in Table \ref{tab:presetting_qa}, pre-setting Q\&A is strictly aligned with the missing slot information.
The average trajectory lengths are 12.74 for the noise tasks and 7.53 for the ambiguous.
Furthermore, we randomly insert one of 150+ ads at a step within one of the 2,504 trajectories in AITZ, ensuring that it overlays the original target button while shifting a step. 
Each trajectory in AITZ-Noisy contains only a single injected advertisement, whereas trajectories in Noisy split include at least five noisy steps, along with additional non-task-related pages such as authorisation, tutorials, redirection pages, and in-app purchase services. 
More details are in Appendix \ref{appendix:casestudy}.

\section{Experiment} 
\subsection{Experimental Setup} \label{experimental setup}
\begin{figure*}[h]
    \centering
    \begin{minipage}[b]{0.3\textwidth}
        \centering
        \includegraphics[width=\textwidth]{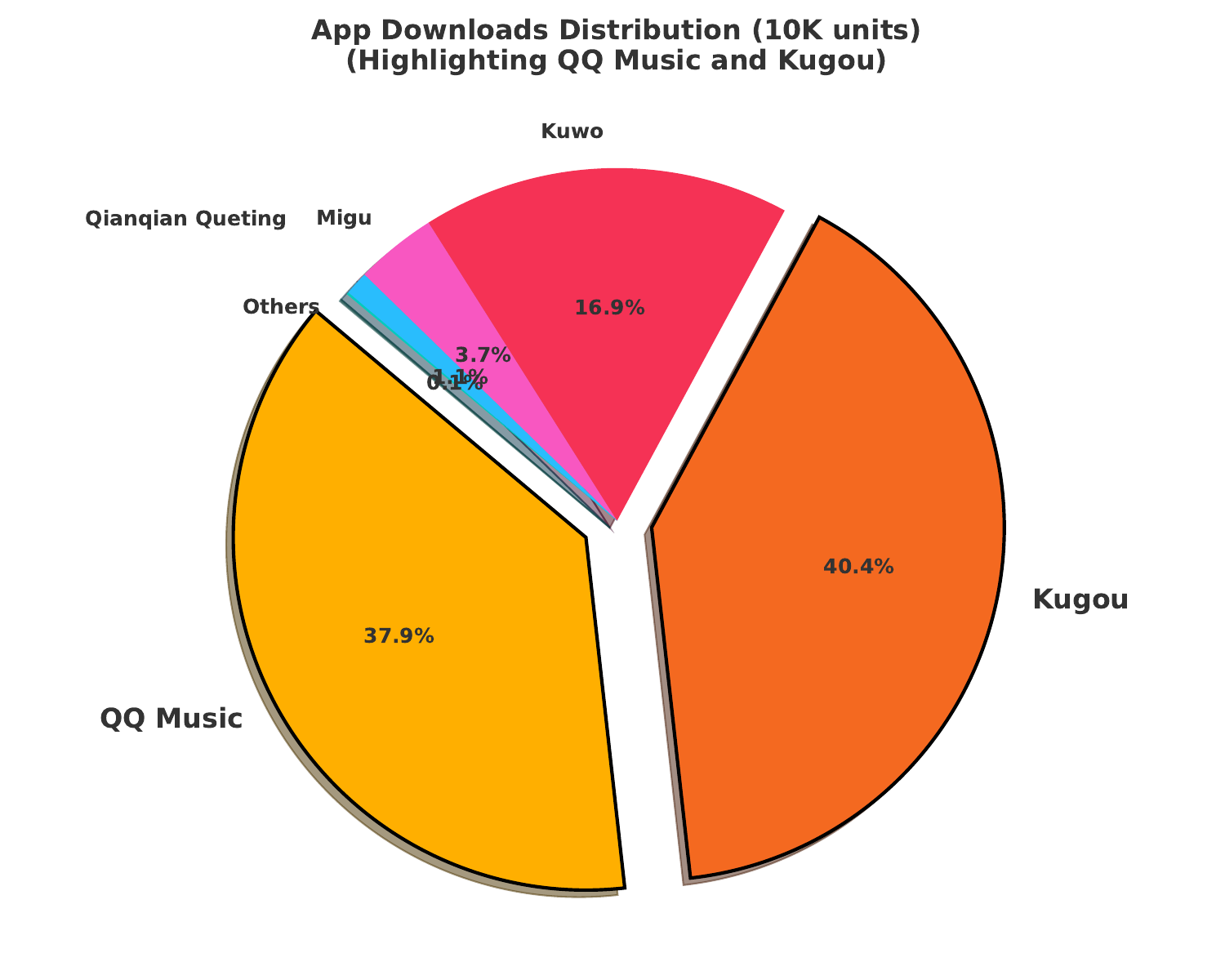}
    \end{minipage}
    \hfill
    \begin{minipage}[b]{0.35\textwidth}
        \centering
        \includegraphics[width=\textwidth]{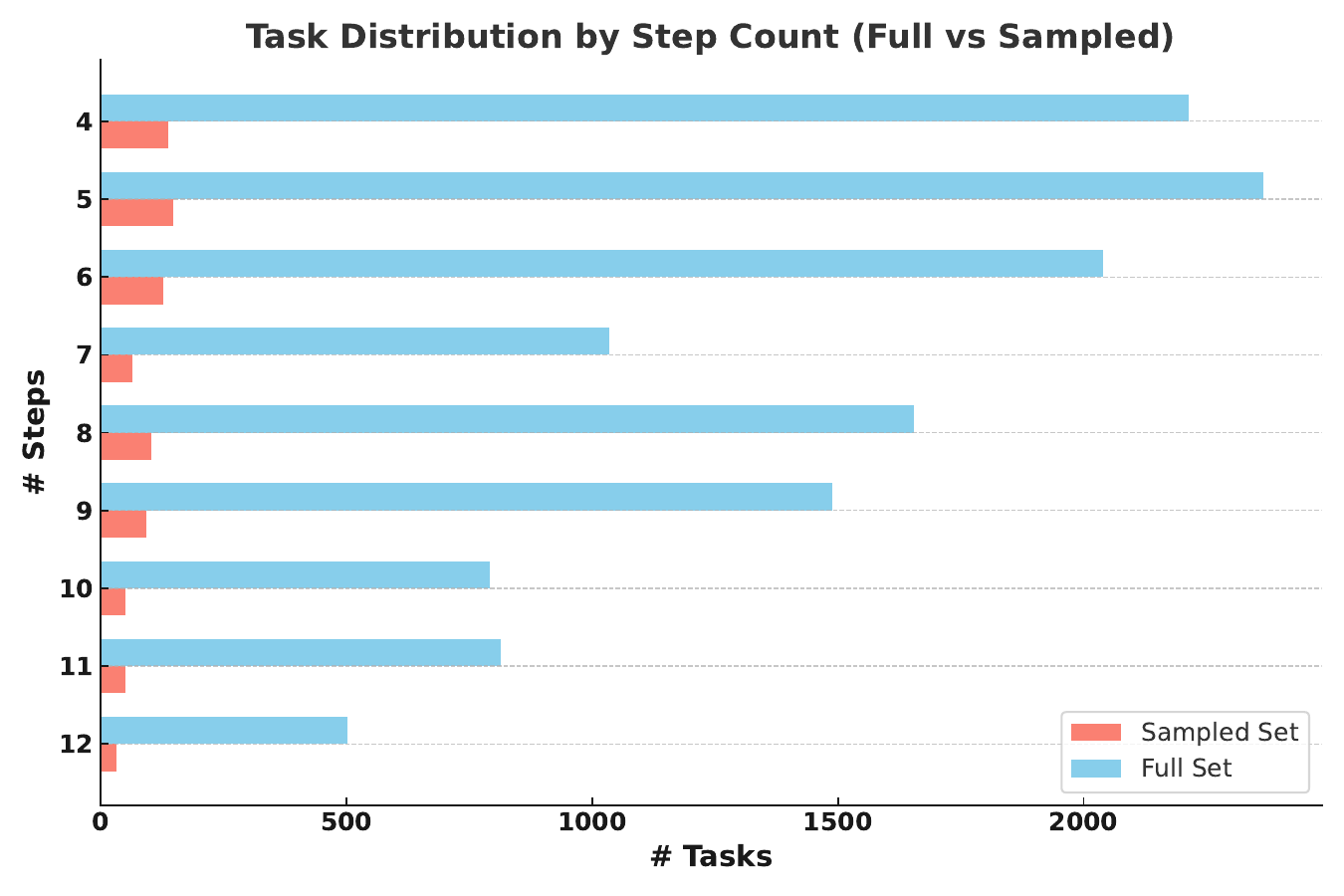}
    \end{minipage}
    \hfill
    \begin{minipage}[b]{0.30\textwidth}
        \centering
        \includegraphics[width=\textwidth]{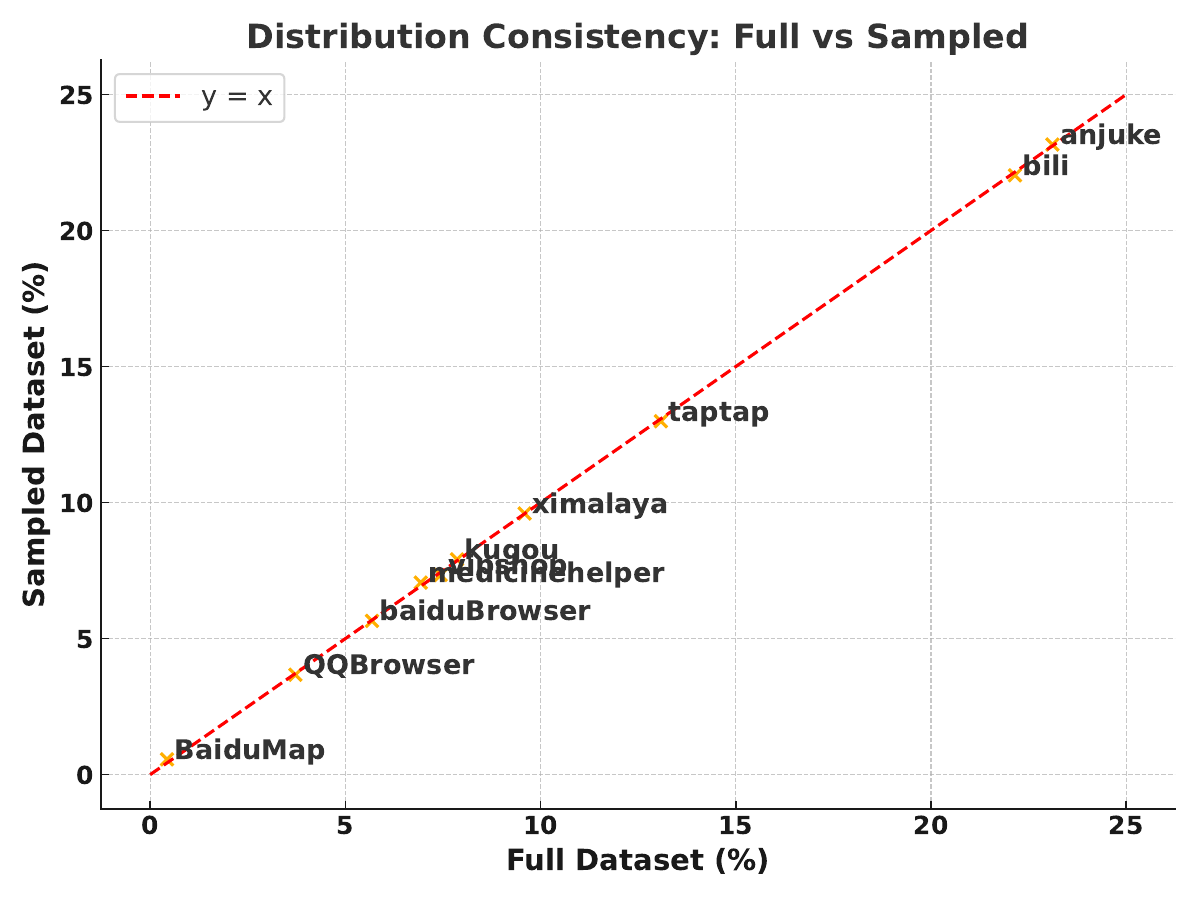}
    \end{minipage}
    \caption{Download volume distribution of music APPs in one month, \textbf{on the left}. The distribution of tasks and their respective step counts, \textbf{in the middle}. The full distribution of apps and their sampled distribution, \textbf{on the right}.}
    \label{fig:app_fan}
\vspace{-10pt}
\end{figure*}

For Mobile-Bench-v2 common, noisy, and ambiguous splits, we experiment agent frameworks such as AppAgent-v1 and MobileAgent-v2 with different fundamental VLMs: InternVL2-40B \citep{chen2024expanding}, LLAVA-72B-NEXT \citep{li2024llava}, Qwen2-VL-72B \citep{wang2024qwen2}, Llama3.2-90B \citep{dubey2024llama}, Qwen-VL-Max, GPT-4o \citep{achiam2023gpt}, and GPT-4v. 
For mobile agents, we use Cogagent \citep{hong2024cogagent}, UGround-7B \citep{gou2024navigating}, OS-Atlas-7B \citep{wu2024atlas}, UI-Tars \citep{qin2025ui}, Kimi-VL \citep{team2025kimi}, and DeepSeek-VL2 \citep{wu2024deepseek}.
To reduce cost, only zero-shot evaluations are done on a subset split of \textit{Random-800}, which has a similar sub-split distribution to the full split. 
As shown in Figure \ref{fig:app_fan}, the distribution of step and app categories in \textit{Random-800} is fully consistent with that of the full dataset, and 800 instructions correspond one-to-one to 800 predefined instruction templates. 
Simple and complex splits set the maximum number of steps to 20 and 25. 
For multi-path, the current page and expectation page are provided for self-reflection, while other actions are restricted to the current one. 
Unlike Appagent, we annotate the widget types using specific numbers (Figure \ref{appendix_2}). 
\subsection{Metrics} 
We establish four metrics proposed in MobileAgentBench \citep{wang2024mobileagentbench} and AUTO-UI \citep{zhang2023you}, 

\noindent \textbf{Success Rate (SR):} $N_{success}/M_{tasks}$, judged by whether the agent reaches the final pages in multi-path evaluation or all actions are correct in single-path evaluation. 

\noindent \textbf{Step Efficiency (SE):} $S_{actual}/S_{min}$, where $S_{actual}$ is the number of actual
steps to complete a task, and $S_{min}$ is the task's minimal annotated steps.
This metric expresses whether the agent performs unnecessary or redundant actions in multi-path evaluation. 

\noindent \textbf{Step Accuracy (Step.Acc):} $S_{tp} / S_{gt}$,  where $S_{tp}$ is the number of predicted actions that match the golden actions in single-path evaluation. This metric also reflects the step reward when the actions are compared with the keynodes in multi-path evaluation.


\noindent \textbf{TYPE Accuracy:} $S_{ttp} / S_{gt}$, where $S_{ttp}$ is the number of predicted actions that match the type of golden actions. For the whole three actions, we use TYPE to check whether the action types are correct. 

\subsection{Data Quality Verification} 
\begin{table}[t]
\centering
\caption{Quality verification results. Crowdsourced annotations verified by agent professionals.}
\scriptsize
\renewcommand{\arraystretch}{0.9}
\begin{tabular}{lcccc}
\toprule
\textbf{Metric} & \textbf{Simple} & \textbf{Complex} & \textbf{Noisy} & \textbf{Ambiguous} \\
\midrule
Annotation Step & 5.62 & 8.21 & 12.74 & 7.53 \\
Evaluation Step & 5.57 & 8.07 & 12.69 & 7.43 \\
\midrule
\textbf{Win Rate$\uparrow$} & 96.0 & 82.5 & 100.0 & 100.0 \\
\textbf{SE$\downarrow$} & 1.01 & 1.02 & 1.00 & 1.01 \\
\bottomrule
\end{tabular}
\label{tab:quality}
\vspace{-0.3cm}
\end{table}
The common split is based on Mobile3M's filtering and annotation, while the noisy and ambiguous instructions are manually annotated. Due to the random walk involved in the former, the instruction trajectories may include redundant actions, whereas the latter requires checking whether the noise is handled correctly and the quality of the Q\&A. 
Therefore, we design a data quality verification experiment shown in Table \ref{tab:quality}. 
Win Rate represents the proportion of instructions generated by GIAS that are equal to or exceed the quality of those manually validated and annotated. 
SE is almost equal to 1, indicating almost no redundant steps in the annotation. 
However, nearly 20\% data constructed in the complex split are found to be suboptimal. 
This is primarily due to more slots in complex instruction templates, which makes unnatural semantics after filling. 
Semantic alignment does not influence invalid steps between the instructions and the actual trajectories. 
Therefore, we only manually revised these instructions to ensure that the \textit{random-800} subset meets the quality requirements.
\begin{table*}[t!]
\caption{Results on MobileBench-v2 Common, Noisy and Ambiguous splits. Type is used in the single-path evaluation, while SE is used in the multi-path evaluation. The ‘-’ indicates that the model does not support Self-Reflection in multi-path evaluation due to not supporting multi-page input.}
\centering
\resizebox{\linewidth}{!}{
\begin{tabular}{l|c|ccc|ccc|ccc|ccc}
    \toprule
    \multirow{2}{*}{\textbf{Models}}& \multirow{2}{*}{\textbf{Cate.}} &\multicolumn{3}{c|}{\textbf{Common-Simple}} &\multicolumn{3}{c|}{\textbf{Common-Complex}} &\multicolumn{3}{c|}{\textbf{Noisy Data}} &\multicolumn{3}{c}{\textbf{Ambiguous Data}}  \\
    &&\textbf{Type/SE} &\textbf{Step. Acc} &\textbf{SR} &\textbf{Type/SE} &\textbf{Step. Acc} &\textbf{SR} &\textbf{Type/SE} &\textbf{Step. Acc} &\textbf{SR} &\textbf{Type/SE} &\textbf{Step. Acc} &\textbf{SR}\\
    \midrule
    \multicolumn{14}{c}{\cellcolor{gray!20} Single-Agent Framework: AppAgent-v1 } \\
        \midrule
    \multirow{2}{*}{InternVL2-40B} & Single	& 82.4 & 35.7 & 1.0 & 85.1& 38.4 & 1.5	& 60.4	& 12.7 & 0.0 & 76.7 & 29.0 & 1.0 \\
    & Multi & 5.8 & 43.0 & 10.1 & 4.4 & 46.0 & 6.5 & - & -	& -	& -	& -	& - \\
    \multirow{2}{*}{LLAVA-72B-NEXT}	& Single & 93.2 & 7.4	& 0.0	& 87.7 & 7.8 & 0.0	& 70.4 & 2.7 & 0.0	& 82.3 & 4.8 & 0.0 \\
    & Multi	& 6.2	& 1.9	& 0.0	& 4.5	& 5.1	& 0.0	& -	& -	& -& 	-	& -	& - \\
    \multirow{2}{*}{Qwen2-VL-72B}  
    & Single & 95.2 & 60.3 & 21.1 & 93.5 & 53.8 & 5.0 & 78.0 & 24.4 & 0.0 & 91.2 & 43.5 & 1.0\\
    & Multi & 5.2 & 62.8 & 20.6 & 4.4 & 58.9 & 4.0 & - & - & -& - & - & -\\
    \multirow{2}{*}{Qwen-VL-Max} 
    & Single & 94.7 & 58.6 & 20.5 & 91.2 & 54.7 & 7.5 & 77.1 & 24.3 & 0.0 & 90.3 & 48.5 &  0.0 \\
    & Multi & 5.9 & 67.6 & 12.6 & 4.3 & 63.1 & 6.6 & - & - & -& - & - & -\\
    \multirow{1}{*}{Llama3.2-VL-90B}  
    & Single & 86.4 & 22.4 & 2.6 & 87.0 & 24.3 & 1.0 & 69.7 & 11.2 & 0.0 & 85.4 & 15.5 & 0.0\\
    \multirow{2}{*}{GPT-4v}  
    & Single & 91.2 & 24.0 & 6.0 & 90.8 & 25.2 & 0.0 & 72.7 & 17.4 & 0.0 & 88.6 & 20.5 & 0.0 \\
    & Multi & 6.1 & 29.7 & 3.0 & 4.5 & 29.4 & 0.0 & - & - & - & - & - & - \\  
    \multirow{2}{*}{GPT-4o} 
    & Single & 80.4 & 57.6 & 18.5 & 79.2 & 50.6 & 11.5  & 72.3 & 18.2 & 0.0 & 94.7 & 33.9 & 0.0 \\
    & Multi & 5.3 & 61.8 & 19.8 & 4.4 & 61.7 & 6.5 & - & - & -& - & - & - \\
    
    \midrule
    \multicolumn{14}{c}{\cellcolor{gray!20} Multi-Agents Framework: MobileAgent-v2} \\
    \midrule
    \multirow{2}{*}{InternVL2-40B} & Single	& 84.5 & 19.3 & 0.0	& 80.7	& 26.4	& 0.5 & 64.3 & 9.1	& 0.0	& 76.0 & 16.3 & 0.0 \\
    & Multi & 6.0 & 27.6 & 3.5	& 4.4 & 32.6 & 3.5 & - & -	& -	& -	& -	& - \\
    \multirow{2}{*}{Qwen2-VL-72B}  
    & Single & 91.5 & 50.5 & 13.0 & 91.6 & 49.0 & 4.5 & 75.8 & 20.7 & 0.0 & 86.2 & 40.8 & 1.0 \\
    & Multi & 5.4 & 54.9 & 15.1 & 4.4 & 58.6 & 4.0 & - & - & - & - & - & -\\
    \multirow{2}{*}{Qwen-VL-Max} 
    & Single & 74.2 & 17.0 & 3.0 & 68.8 & 12.3 & 2.0 & 66.5 & 4.2 & 0.0 & 66.6 & 9.0 & 0.0 \\
    & Multi & 5.4 & 29.6 & 4.5 & 4.3 & 24.8 & 3.0 & - & - & - & - & - & - \\
    \multirow{1}{*}{Llama3.2-VL-90B}  
    & Single & 62.4 & 16.6 & 1.0 & 67.0 & 17.5 & 0.0 & 63.7 & 9.7 & 0.0 & 64.3 & 8.3 & 0.0\\
    \multirow{2}{*}{GPT-4v}  
    & Single & 90.8 & 22.9 & 3.8 & 90.6 & 28.3 & 0.5 & 62.5 & 12.6 & 0.0 & 91.0 & 15.6 & 0.0 \\
    & Multi & 6.0 & 17.8 & 5.4 & 4.5 & 11.8 & 0.0 & - & - & - & - & - & -\\
    \multirow{2}{*}{GPT-4o} 
    & Single & 91.9 & 53.5 & 13.5 & 92.3 & 50.5 & 17.0 & 77.1 & 25.5 & 2.0 & 91.6 & 39.7 & 12.0\\
    & Multi & 4.9 & 57.6 & 25.6 & 4.2 & 56.3 & 17.5 & - & - & - & - & - & -\\
    \midrule
    \multicolumn{14}{c}{\cellcolor{gray!20} Open-sourced Mobile Agents} \\
    \midrule
    CogAgent-18B & Single & 75.6 & 20.9 & 13.0 & 62.3 & 20.8 & 6.0 & 57.2& 16.2 & 1.0 & 72.4 & 30.6 & 11.0 \\
    UGround-7B & Single & 73.0 & 39.5 & 17.0 & 73.8 & 36.0 & 10.0 & 71.2 & 32.8 & 2.0 & 79.9 & 47.0 & 20.0 \\
    UI-Tars-7B-dpo & Single & 75.3 & 41.8 & 19.0 & 75.9 & 37.8 & 12.0 & 73.2 & 35.6 & 4.0 & 81.8 & 49.4 & 23.0 \\
    OS-Atlas-7B-pro & Single & 82.1 & 51.5 & 28.0 & 83.5 & 50.6 & 18.0 & 81.2 & 45.3 & 2.0 & 86.4 & 52.3 & 24.0 \\
    Kimi-VL-A3B & Single & 75.1 & 21.7 & 12.0 & 61.8 & 21.5 & 6.5 & 56.9 & 15.8 & 1.0 & 71.7 & 29.9 & 11.0 \\
    DeepSeek-VL2 & Single & 72.6 & 38.8 & 16.0 & 73.1 & 35.2 & 9.5 & 70.5 & 32.1 & 2.0 & 79.2 & 46.1 & 19.0 \\
    UI-Tars-72B-dpo & Single & 94.3 & 64.2 & 32.0 & 96.0 & 63.5 & 24.0 & 94.5 & 59.8 & 7.0 & 92.5 & 66.0 & 30.0 \\
    \bottomrule
\end{tabular}
 }
\label{tab:main result}
\vspace{-12pt}
\end{table*}

\subsection{Main Result}

\noindent \textbf{Common data results.} 
As shown in Table \ref{tab:main result}, in terms of \textbf{agent framework differences}, AppAgent-v1 outperforms in single-path evaluations, whereas MobileAgent-v2 excels in multi-path scenarios. 
This is because single-path evaluation presets the correct historical action and guides agents focus on the current page. 
In multi-path evaluation, agents' actions depend on their previous decision and pages, so agents are more prone to being trapped in mistaken execution trajectories.
MobileAgent-v2's reflection and expectation mechanism effectively mitigates this problem. 
From the perspective of the \textbf{foundation model}, Qwen and GPT series models show superior performance in commom tasks, while others perform relatively poorly (50-70\% compared to 20-40\%). 
Due to the weaker instruction-following and grounding capabilities of smaller models, their overall performance is limited. The differences in planning ability across models are relatively minor, as reflected in the comparable TYPE accuracy. 
Open-source mobile agents exhibit substantial potential compared to mobile framework VLMs. Notably, OS-Atlas-7B achieves performance on par with GPT-4o, and UI-Tars-72B achieves the best results across all metrics.
However, all these agents demonstrate a common limitation: even when the action type is correctly planned, they often fail to select or ground the correct UI element (TYPE $\gg$ Step.Acc). 
Considering the evaluation method, the consistently higher performance observed under multi-path evaluation compared to single-path evaluation highlights the variability in trajectory choices across agents. 
Although the step-rewards are comparable, InternVL takes over five times more steps to get them compared to GPT-4o (SE 24.7 \textit{vs.} 4.4).
This observation proves that muilti-path evaluation can offer a more faithful assessment of model effectiveness.

\begin{table}[t]
\centering
\caption{Ablation study on \textbf{Mobile-Bench-Ambiguous}: comparison of AppAgent-v1 and MobileAgent-v2 under full and ambiguous instructions.}
\scriptsize
\renewcommand{\arraystretch}{1.1}
\begin{tabular}{l|cc|cc|cc|cc}
\toprule
\multirow{2}{*}{\textbf{Model}} 
& \multicolumn{2}{c|}{\textbf{AppAgent (Full)}} 
& \multicolumn{2}{c|}{\textbf{AppAgent (Ambig.)}} 
& \multicolumn{2}{c|}{\textbf{MobileAgent (Full)}} 
& \multicolumn{2}{c}{\textbf{MobileAgent (Ambig.)}} \\
& Type & StepAcc & Type & StepAcc & Type & StepAcc & Type & StepAcc \\
\midrule
InternVL2-40B & 70.8 & 21.6 & 76.7 (+5.9) & 29.0 (+7.4) & 61.5 & 8.3 & 76.0 (+15.5) & 16.3 (+8.0) \\
Qwen2-VL-72B  & 82.9 & 41.5 & 91.2 (+8.3) & 43.5 (+2.0) & 80.3 & 38.5 & 86.2 (+5.9) & 40.8 (+2.3) \\
Qwen-VL-Max   & 81.2 & 39.3 & 90.3 (+9.1) & 48.5 (+9.2) & 57.6 & 3.3 & 66.6 (+9) & 9.0 (+5.7) \\
Llama3.2-VL-90B & 67.9 & 7.6 & 85.4 (+17.5) & 15.5 (+7.9) & 57.8 & 4.0 & 64.3 (+6.5) & 8.3 (+4.3) \\
GPT-4v        & 72.8 & 15.9 & 88.6 (+15.8) & 20.5 (+4.6) & 84.6 & 13.9 & 91.0 (+6.4) & 15.6 (+1.7) \\
GPT-4o        & 79.7 & 31.9 & 94.7 (+15) & 33.9 (+2) & 87.7 & 38.4 & 91.6 (+3.9) & 39.7 (+1.3) \\
\bottomrule
\end{tabular}
\label{tab:ambiguous_result}
\vspace{-10pt}
\end{table}

\noindent \textbf{Out-domain Mobile-Bench-Noisy results.} 
For noisy data, the results in the third block of Table \ref{tab:main result} demonstrate that only a few VLMs complete a task. 
All VLMs exhibit a declining trend in Step.Acc, particularly LLAVA and Qwen2-VL, while this phenomenon is also observed in open-source agents, due to the absence of noise in their training data.
Unlike AITZ-Noise, the ads in Noisy-App are more dynamic and variable which as shown in Figure \ref{noisy_mobile}. 
Specifically, these noises exhibit the following three features: (1) After the pop-up ad countdown ends, the ad disappears automatically, and the agent's delayed instructions may cause accidental taps; (2) Some video ads cannot be closed during the early viewing stages; (3) The mis-taps caused by real ad noise may trigger app redirection.

\begin{table*}[ht]
\caption{Results on AITZ-Noise. Qwen2-VL and OS-Atlas are evaluated on AITZ and AITZ-Noise.}
\centering
\resizebox{\linewidth}{!}{
\renewcommand{\arraystretch}{1.2}
\begin{tabular}{lcccccccccccc}
\hline\hline
\multirow{2}{*}{\textbf{Agent}} & \multirow{2}{*}{\textbf{Benchmark}} & \multicolumn{2}{c}{\textbf{General}} & \multicolumn{2}{c}{\textbf{Google App}} & \multicolumn{2}{c}{\textbf{Install}} & \multicolumn{2}{c}{\textbf{Web Shopping}} & \multicolumn{2}{c}{\textbf{Total}}\\ 
& &  \textbf{Step. Acc} & \textbf{Noisy} & \textbf{Step. Acc} & \textbf{Noisy} & \textbf{Step. Acc} & \textbf{Noisy} & \textbf{Step. Acc} & \textbf{Noisy} & \textbf{Step. Acc} & \textbf{Noisy} \\ \hline 
\multirow{2}{*}{\textbf{Qwen2-VL-7B}} & Normal & 38.5 & - & 44.8 & - & 60.0 & - & 45.1 & - & 46.9 \\
& Noise & 37.3 & 15.4 & 42.2 & 17.2 & 54.7 & 20.5 & 42.1 & 17.1 & 43.9 & 17.4 \\
\multirow{2}{*}{\textbf{OS-Atlas-7B}} & Normal & 41.9 & -& 46.4 & -& 60.5 & - & 46.3 & - & 48.6 \\
& Noise & 38.8 & 21.8 & 41.7 & 19.7 & 56.4 & 23.5 & 43.8 & 23.6  & 45.1 & 21.7 \\
\hline \hline
\end{tabular}
}
\label{tab:Noisy_data_results1}
\vspace{-0.3cm}
\end{table*}

\noindent \textbf{Out-domain AITZ-noisy results.} 
As shown in Table \ref{tab:Noisy_data_results1}, the Step.Acc of Qwen2-VL and OS-Atlas decreased by an average of 3.0\% and 3.5\% from AITZ normal to in-domain AITZ-Noisy. 
Given that the Noisy step accuracy is 17.4\% and 21.7\%, this indicates that open-source agents fail to learn the features of advertisements because a few ads still exist in their training data. 
They exhibit almost no generalization capability on transferred noisy data, even when only the background screenshot changes. 
When agents become trapped in a page unrelated to the current task, they struggle to determine how to proceed next.
More details can be found in Appendix \ref{DataContamination}. 

\subsection{Ambiguous Instruction Ablation Study}
As shown in the far-right column of Table \ref{tab:ambiguous_result}, all agents exhibit improved performance, even 17.5\%, when supplied with more informative context through step-by-step Q\&A.
Ablation results demonstrate that the active interaction module can help agents effectively ignore irrelevant content in task instructions for the current step. 
This is because full instructions may affect the agent's ability to identify tasks on the current page accurately. 
In contrast, ambiguous instructions with step-by-step Q\&A help the agent better comprehend the page and execute more appropriate actions. 
However, it is worth noting that not all agents benefit from this mechanism. 
Weaker VLMs (e.g., InternVL, +5.9\%) struggle to generate effective questions, while stronger VLMs (e.g., GPT-4o, +3.9\%) are already capable of effective planning and decision-making without additional support. VLMs with intermediate performance, such as LLaMA3.2-VL-90B, benefit more from this mechanism (+17.5\%).
\subsection{Discussion}
\begin{table*}[ht]
\caption{Qwen2-VL, Cogagent, and OS-Atlas evaluated on AITZ-Noise. Metric ``Noisy'' means in-domain noisy step accuracy. More Experiments can be seen in Table \ref{tab:Noisy_data_results3}.}
\centering
\resizebox{\textwidth}{!}{
\begin{tabular}{lp{2.2cm}|ccc|ccc|ccc|ccc}
\toprule
\multirow{2}{*}{\textbf{Agent}} & \multirow{2}{*}{\textbf{Training Data}} & \multicolumn{3}{c|}{\textbf{General}} & \multicolumn{3}{c|}{\textbf{Google App}} & \multicolumn{3}{c|}{\textbf{Install}} & \multicolumn{3}{c}{\textbf{Web Shopping}} \\ 
& & \textbf{Step. Acc} & \textbf{Noisy} & \textbf{SR} & \textbf{Step. Acc} & \textbf{Noisy} & \textbf{SR} & \textbf{Step. Acc} & \textbf{Noisy} & \textbf{SR} & \textbf{Step. Acc} & \textbf{Noisy} & \textbf{SR} \\ 
\midrule
\multicolumn{14}{c}{\cellcolor{gray!20} Supervised Fine-tuning Setting(LoRA)} \\
\midrule
CogAgent-18B & AITW(CoaT) & 40.4 & - & 11.5 & 38.1 & - & 11.3 & 45.2 & - & 17.3 & 39.1 & - & 13.4 \\ 
Qwen2-VL-7B & AITZ(CoaT) & 36.1 & - & 8.3 & 39.1 & - & 11.2 & 50.9 & - & 20.7 & 41.8 & - & 15.2 \\ 
Qwen2-VL-7B & AITZ-Noise & 39.8 & 98.0 & 11.7 & 42.3 & 99.0 & 16.6 & 60.9 & 100 & 30.4 & 41.5 & 99.0 & 13.3 \\ 
OS-Atlas-7B & AITZ-Noise & 46.2 & 99.0 & 18.7 & 50.2 & 99.5 & 21.3 & 62.4 & 100 & 33.0 &
44.8 & 99.0 & 17.3 \\
\midrule
\multicolumn{14}{c}{\cellcolor{gray!20} Supervised Fine-tuning Setting(Full)} \\
\midrule
Qwen2-VL-7B &  AITZ-Noise & 43.2 & 96.0 & 15.6 & 46.2 & 97.5 & 19.8 & 64.2 & 98.5 & 35.6 & 50.9 & 98.0 & 22.3 \\ 
OS-Atlas-7B & AITZ-Noise & 47.2 & 98.0 & 19.0 & 47.1 & 99.0 & 22.3 & 66.7 & 99.0 & 38.0 &
51.8 & 99.0 & 23.5 \\
\toprule
\end{tabular}
}
\label{tab:Noisy_data_results2}
\end{table*}
\noindent \textbf{Solving in-domain noise through post-training.} 
We are more focused on whether increasing the proportion of noisy training data can address the in-domain noisy problem. As shown in Table \ref{tab:Noisy_data_results2}, Qwen2-VL, compared to the original AITZ training data, shows a Step.Acc improvement of 3.7\%, 3.2\%, 10.0\%, -0.3\% and SR improvement of 3.4\%, 5.4\%, 9.7\%, -1.9\% on the four sub-tasks. 
At the same time, full parameter fine-tuning outperforms LoRA in overall results but performs slightly worse than LoRA on noise step processing (an average of 1.5\% lower). 
After training, the agent is able to correctly handle the vast majority of noisy steps (with an accuracy greater than 97\%), demonstrating the effectiveness of training with noisy data.
\section{Conclusion}
In this paper, we propose Mobile-Bench-v2, a more realistic and comprehensive mobile agent benchmark that includes common instruction trajectories, noisy app split, noisy contaminated split, and ambiguous instruction split. We also propose a novel slot-based trajectory annotation method without human evaluation, named GIAS and an offline multi-path evaluation method. 
This benchmark provides a foundation for evaluating and optimizing GUI agent studies focused on multi-path solutions searching, noise robustness, and proactive interaction.

\newpage
\bibliographystyle{unsrt}
\bibliography{mobile-bench-v2}
\newpage
\appendix
\section{Mobile3M Dataset}\label{appendix:randomwalk}
\begin{figure*}[!t]
  \centering
  \includegraphics[width=1\textwidth]{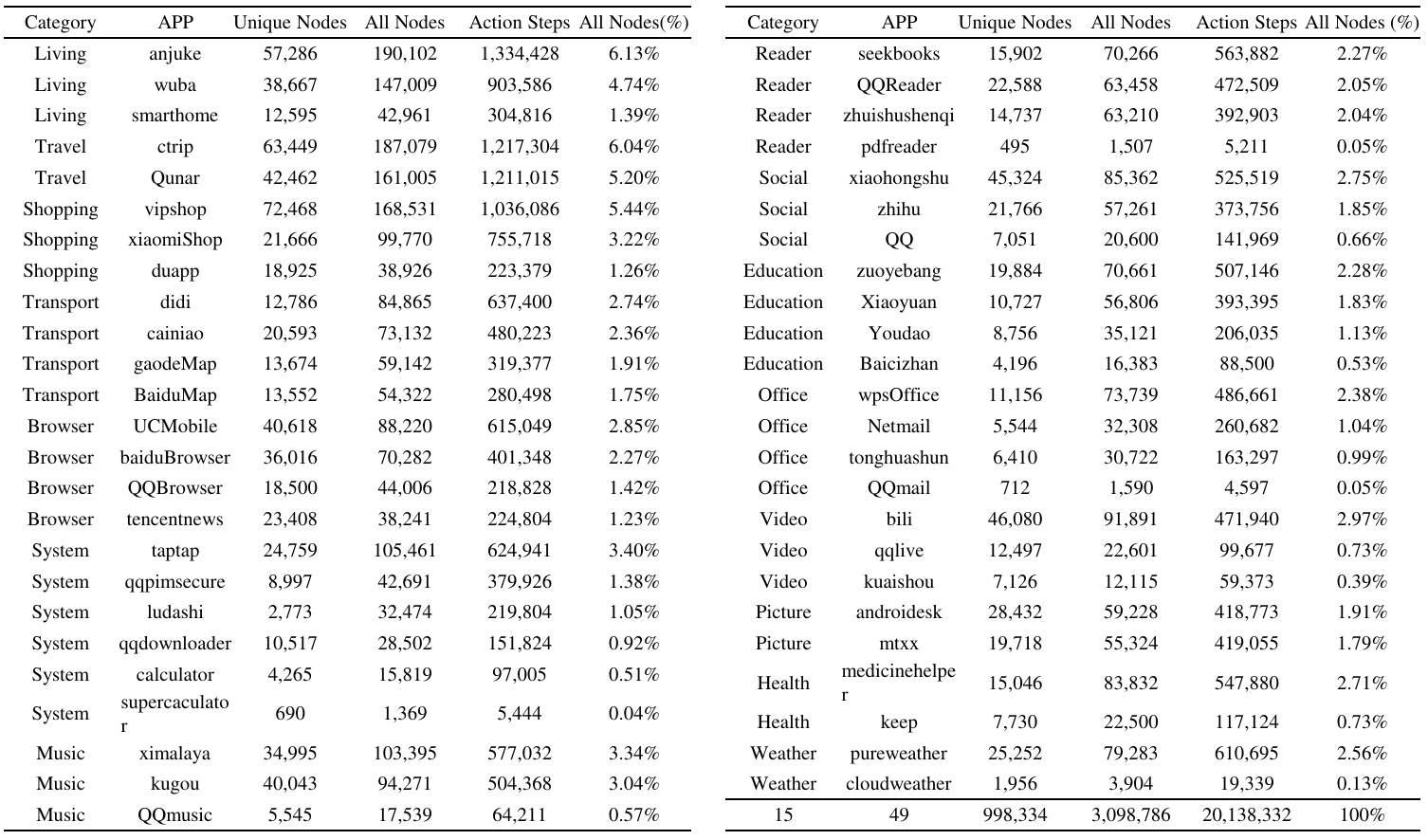}
  \caption{ The data distribution in Mobile3M.}
  \label{mobile3m_count}
\end{figure*}
\label{sec:appendix}
Mobile3M is a large-scale dataset designed to systematically explore and analyze the functionality of mobile applications through UI-based interactions. It provides a comprehensive representation of user interface (UI) elements, interactions, and app navigation patterns. Mobile3M is characterized by the following key features:

    (1) \textbf{Scale and Diversity} \\
    Mobile3M includes over 20 million user interactions, covering 3 million screenshots and corresponding XML documents. These data are organized into directed graphs for 49 widely-used Chinese apps, where nodes represent UI pages, and edges capture user actions.

    (2) \textbf{Detailed UI Representation} \\
    Each UI page is described by both a screenshot and an XML document. The XML documents provide detailed structural information, including UI elements (e.g., buttons, text fields), their hierarchical relationships, and layout properties such as bounding boxes.

    (3) \textbf{Action Space} \\
    The dataset defines three fundamental user actions---\textit{click}, \textit{scroll}, and \textit{input}---to simulate real-world app interactions. Each UI page contains an action space derived from its interactable elements, facilitating comprehensive modeling of user behaviors.

    (4) \textbf{Graph-Based Organization} \\
    Mobile3M employs a breadth-first search (BFS) algorithm to explore app functionality, representing the exploration results as graphs. This structure enables the identification of app workflows, the relationship between UI pages, and the possible transitions triggered by user actions.

    (5) \textbf{Efficiency and Optimization} \\
    To enhance exploration efficiency, Mobile3M incorporates a ``unique page'' mechanism that eliminates duplicates by comparing UI pages using a combination of element and pixel-based similarity thresholds. This reduces the exploration space, prevents redundant actions, and avoids cyclic sequences, ensuring more diverse and meaningful data coverage.

    (6) \textbf{Balanced Action Distribution} \\
    The dataset emphasizes balanced representation of user actions by prioritizing underrepresented interactions, such as \textit{input}. For example, random keywords are introduced for input actions, and scroll actions are executed in multiple directions to capture diverse app behaviors.

    (7) \textbf{Task-Oriented Exploration} \\
    Inspired by APPAgent, the dataset leverages a random walk algorithm to systematically interact with UI elements and record transitions between pages. The exploration process captures action traces, enabling task-driven navigation and detailed understanding of app functionalities.
\section{Data Analysis and Construction}\label{appendix:data}
\subsection{GIAS Prompt}

\textbf{Input Format}

\textit{You will be provided with a series of user interaction histories, each consisting of a caption describing the current page and an action performed by the user.}

\textbf{Your Task}

\textit{Analyze each action and the corresponding page caption to determine what action was taken on that page. Summarize these actions into a task description, which should be a request. For example:}
\begin{itemize}
    \item \textit{``I want to see what VIP privileges are available.''}
    \item \textit{``Help me find pants on sale.''}
    \item \textit{``Tell me what items are in my shopping cart.''}
\end{itemize}

\noindent\textbf{Important Notes:}
\begin{enumerate}
    \item \textit{The task description and the sequence of actions should have a logical relationship.}
    \item \textit{The task description should be phrased as a request, reflecting the goal of the actions taken.}
    \item \textit{Actions and captions should be analyzed in sequence to deduce the user's objective.}
\end{enumerate}

\textbf{Output Format}

\textbf{``step-by-step description''}: \textit{``Provide a series of interactions, where each entry corresponds to a screenshot caption of the current phone screen and the action performed on that page.''}

\textbf{``concise task''}: \textit{``Summarize the user's overall goal based on the step-by-step description.''}

\textbf{Example}

\noindent\textbf{Caption 1:}\\
\textit{This image shows a screenshot of a shopping application interface.}

\noindent\textbf{Action 1:}\\
\textit{Click(Skincare Set)}

\noindent\textbf{Caption 2:}\\
\textit{This image shows a screenshot of a shopping application interface. At the top, there is a search bar with the text “Skincare Set.” Additionally, at the bottom of the page, there is a navigation bar with options like ``All Products,'' ``New Arrivals,'' ``Moisturizing,'' ``Dry Skin,'' ``Niacinamide,'' and ``Hyaluronic Acid.'' The current state is ``All Products.''}

\noindent\textbf{Action 2:}\\
\textit{Click(New Arrivals)}

\noindent\textbf{Caption 3:}\\
\textit{This image shows a screenshot of a shopping application interface. At the top, there is a search bar with the text ``Skincare Set.'' Additionally, there is a navigation bar at the bottom with options like ``All Products,'' ``New Arrivals,'' ``Moisturizing,'' ``Dry Skin,'' ``Niacinamide,'' and ``Hyaluronic Acid.'' The current state is ``New Arrivals.'' Below are multiple product recommendations.}

\noindent\textbf{Action 3:}\\
\textit{Click(Ad)}

\noindent\textbf{Caption 4:}\\
\textit{This image shows a product detail page. At the top, there is a pink banner that reads ``Buy a set and get 13 items free,'' along with a product photo.}

\noindent\textbf{Output:}

\noindent\textbf{``step-by-step description'':}
\begin{enumerate}[label=\arabic*.]
    \item \textit{Click the ``Skincare Set'' product under the ``Beauty'' subcategory of ``Recommended.''}
    \item \textit{On the Skincare Set search results page, click the ``New Arrivals'' tab.}
    \item \textit{On the product details page, click the ``Ad'' tab.}
\end{enumerate}

\noindent\textbf{``Concise task'':}\\
\textit{Help me find the latest skincare set that is on promotion.}

\textbf{New Input and Task}

\textit{Now, based on the following input, please generate the ``step-by-step description'' and ``concise task'':}

\noindent\texttt{\{trajectory\_description\}}

\subsection{Data Contamination}\label{DataContamination}
\begin{figure*}[!t]
  \centering
  \includegraphics[width=1\textwidth]{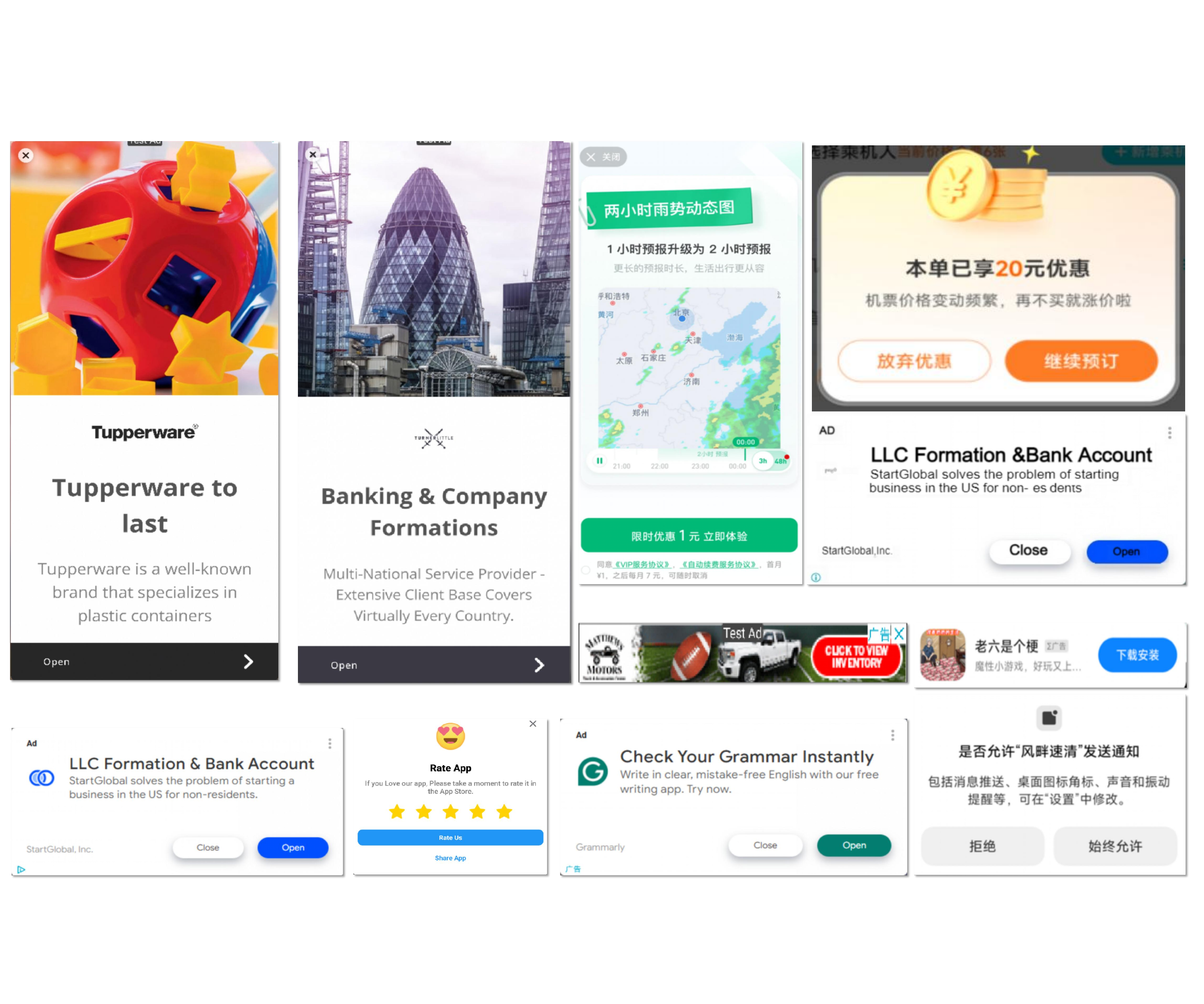}
  \caption{A collection of pop-up ads, collected from Google service official apps, third-party market apps, and mobile apps in mainland China. }
  \label{ads}
\end{figure*}

The collected advertisements are shown in Figure \ref{ads}. We embed them into the normal dataset and applied background whitening. We ensure that the elements that should have been clicked on the current page are no longer visible after the contamination. 
When splitting the training and test data, the position of the embedded advertisements is randomly assigned. However, the types of advertisements in the training data are largely consistent with those in the test data, and the same advertisements maintain consistent embedding positions. 
\begin{table*}[ht]
\centering
\caption{Qwen2-VL and OS-Atlas fine-tuned on AITZ-Noise, AITW, or AITZ and evaluation on AITZ-Noise(Out-domain). Metric ``Noisy'' means out-domain noisy step accuracy.}
\resizebox{\textwidth}{!}{
\begin{tabular}{lp{2.2cm}|cc|cc|cc|cc|cc}
\toprule
\multirow{2}{*}{\textbf{Agent}} & \multirow{2}{*}{\textbf{Training Data}} & \multicolumn{2}{c|}{\textbf{General}} & \multicolumn{2}{c|}{\textbf{Google App}} & \multicolumn{2}{c|}{\textbf{Install}} & \multicolumn{2}{c}{\textbf{Web Shopping}} 
& \multicolumn{2}{c}{\textbf{Total}}\\ 
& & \textbf{Step. Acc} & \textbf{Noisy} & \textbf{Step. Acc} & \textbf{Noisy} & \textbf{Step. Acc} & \textbf{Noisy} & \textbf{Step. Acc} & \textbf{Noisy} & \textbf{Step. Acc} & \textbf{Noisy} \\ 
\midrule
\multicolumn{12}{c}{\cellcolor{gray!20} Normal Data Supervised Fine-tuning} \\
\midrule
Qwen2-VL-7B &  AITZ & 37.33 & 15.38 & 42.18 & 17.11 & 54.68 & 20.45 & 42.06 & 17.14 & 43.84 & 17.46 \\
OS-Atlas & 
AITZ & 38.81 & 21.79 & 41.61 & 19.73 & 56.37 & 23.57 & 43.71 & 23.57 & 45.16 & 21.62 \\
\midrule
\multicolumn{12}{c}{\cellcolor{gray!20} In-domain Noise Supervised Fine-tuning} \\
\midrule
Qwen2VL & AITZ + Noisy & 43.07 & 77.56 & 47.63 & 73.68 & 60.64 & 75.76 & 44.02
& 75.00 & 48.20 & 75.79 \\
OS-Atlas & AITZ + Noisy & 44.92 & 82.05 & 49.64 & 76.32 & 63.06 & 79.55 &
48.01 & 78.57 & 50.99 &79.56 \\
\midrule
\multicolumn{12}{c}{\cellcolor{gray!20} Out-domain Noise Supervised Fine-tuning} \\
\midrule
Qwen2VL & AITZ + Noisy & 37.18 &50.64 &45.18 &41.89 &57.45 &53.38 &42.32 & 50.00 &44.96 &49.90 \\
OS-Atlas & AITZ + Noisy & 41.75 & 53.85 & 45.47 & 48.65 & 60.27 & 60.90 &
47.28 & 55.71 & 48.69 & 55.47 \\
\toprule
\end{tabular}
}
\label{tab:Noisy_data_results3}
\end{table*}
\begin{figure*}[!t]
  \centering
  \includegraphics[width=1\textwidth]{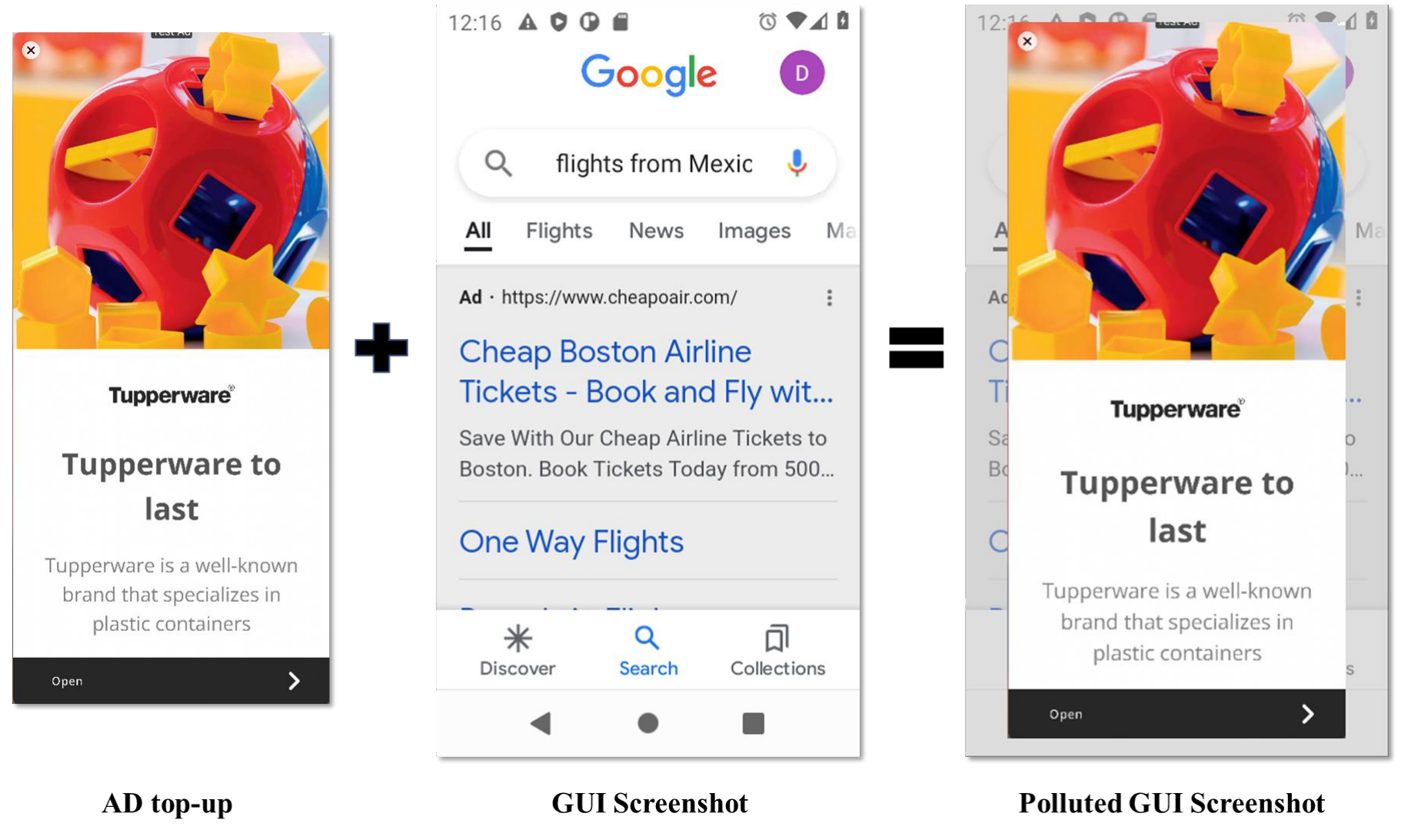}
  \caption{Contaminated datasets are constructed by inserting advertisements and whitening the background of the original GUI screenshots.}
  \label{datapollution}
\end{figure*}
\begin{figure*}[!t]
  \centering
  \includegraphics[width=1\textwidth]{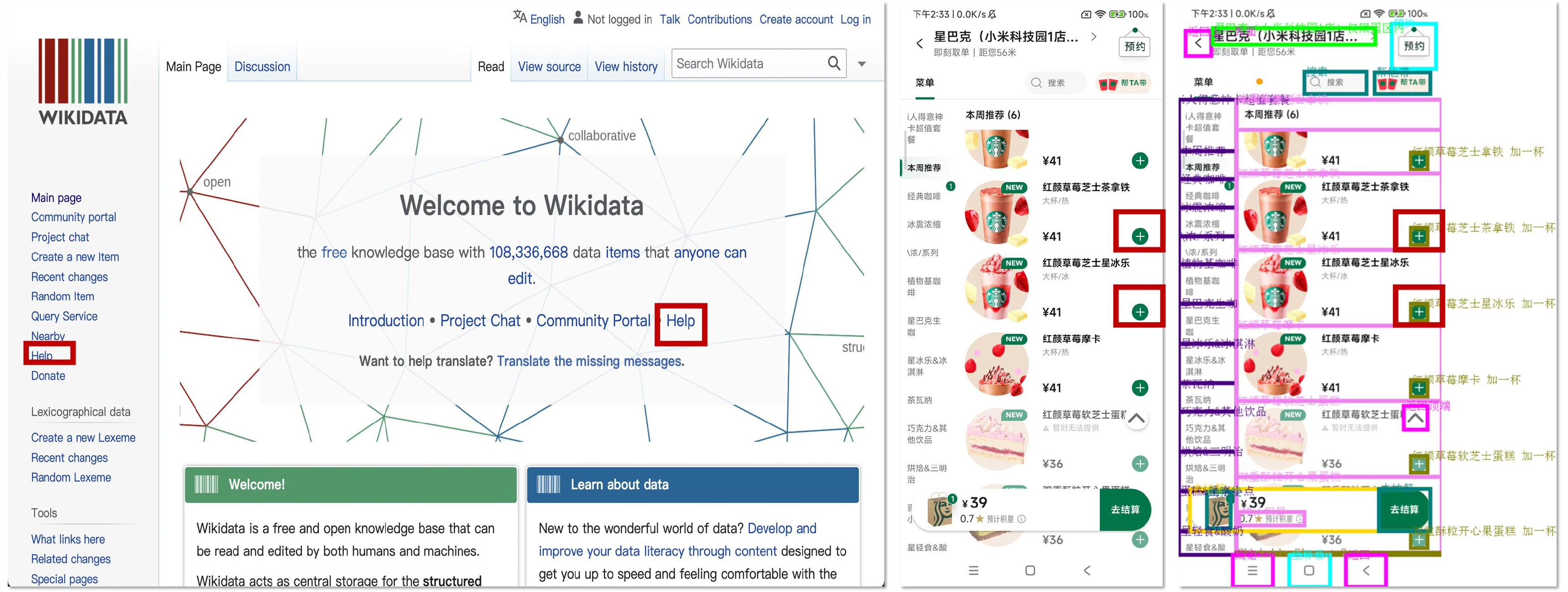}
  \caption{The red boxed areas represent identical graphical controls and identical textual controls, which can create ambiguity in the action history.}
  \label{same_button}
\end{figure*}

\section{Experiment Details} \label{experimental details}
\subsection{Baseline Model Demonstration}\label{appendix:baseline model}
\begin{figure*}[!t]
  \centering
  \includegraphics[width=0.85\textwidth]{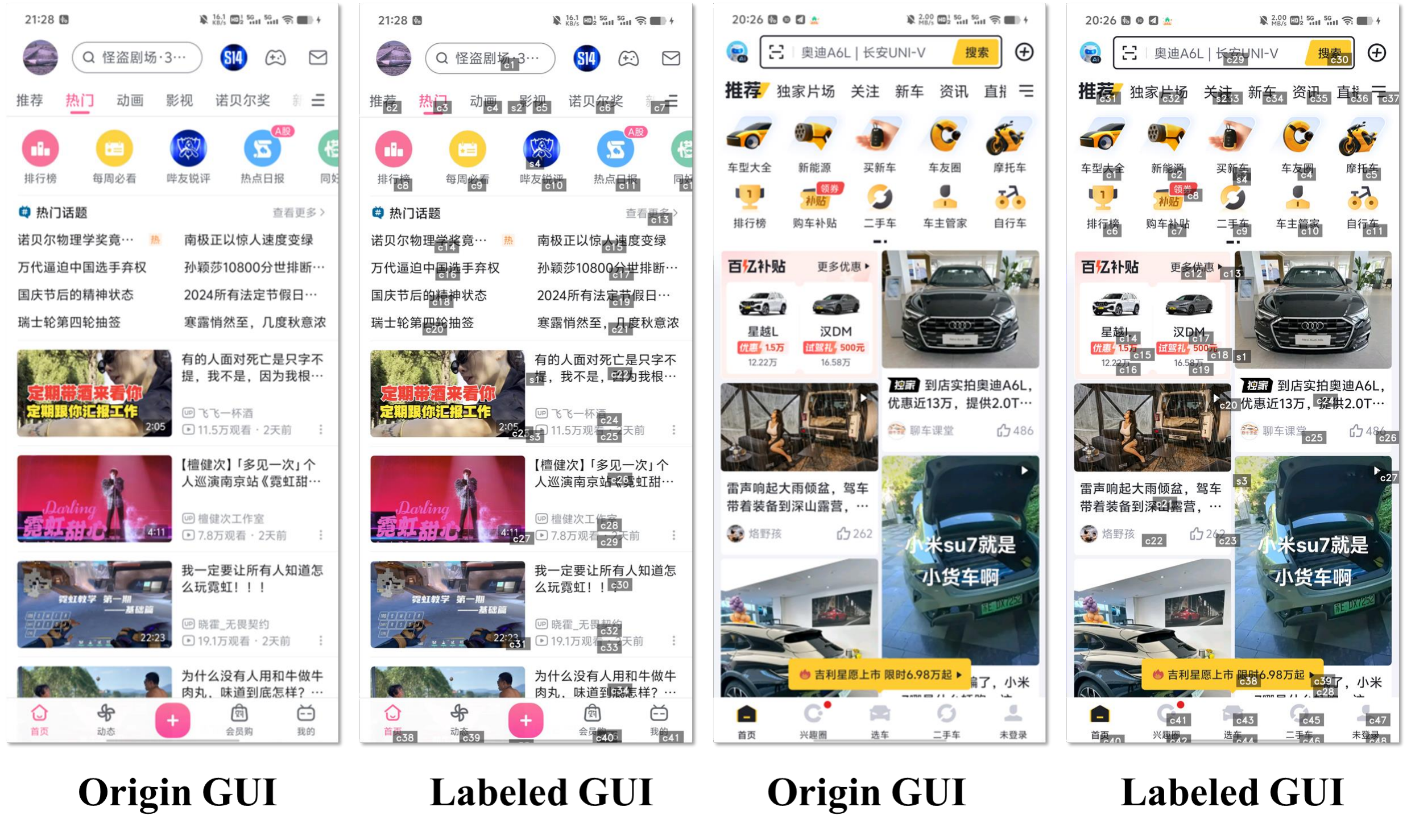}
  \caption{Appagent GUI labeled method. }
  \label{appendix_2}
\end{figure*}

\textbf{AppAgent}
Below is the prompt we used. We did not re-adapt or adjust the prompt for different base models to ensure fairness. 
\begin{lstlisting}[caption={Click Document Template}]
I will give you the screenshot of a mobile app, the clickable UI element is labeled 
with a letter 'c' and the number <ui_element> on the screen. The tag of each element is located at the center of the 
element. Clicking on this UI element is a necessary part of proceeding with a larger task, which is to <task_description>.
In order to realize this larger task, you must first realize the current task <current_task_desc> in current screenshot.
Your task is to describe the functionality of the UI element concisely in one or two sentences. Notice that your 
description of the UI element should focus on the general function. For example, if the UI element is used to navigate 
to the chat window with John, your description should not include the name of the specific person. Just say: 
"Clicking this area will navigate the user to the chat window". Never include the tag of the
UI element in your description. You can use pronouns such as "the UI element" to refer to the element.
\end{lstlisting}

\begin{lstlisting}[caption={Click Documentation Template}]
I will give you the screenshot of a mobile app, the clickable UI element is labeled 
with a letter 'c' and the number <ui_element> on the screen. The tag of each element is located at the center of the 
element. Clicking on this UI element is a necessary part of proceeding with a larger task, which is to <task_description>.
In order to realize this larger task, you must first realize the current task <current_task_desc> in current screenshot.
Your task is to describe the functionality of the UI element concisely in one or two sentences. Notice that your 
description of the UI element should focus on the general function. For example, if the UI element is used to navigate 
to the chat window with John, your description should not include the name of the specific person. Just say: 
"Clicking this area will navigate the user to the chat window". Never include the tag of the
UI element in your description. You can use pronouns such as "the UI element" to refer to the element.
\end{lstlisting}

\begin{lstlisting}[caption={Refine Documentation Suffix}]
A documentation of this UI element generated from previous demos is shown below. Your 
generated description should be based on this previous doc and optimize it. Notice that it is possible that your 
understanding of the function of the UI element derived from the given screenshots conflicts with the previous doc, 
because the function of a UI element can be flexible. In this case, your generated description should combine both.
Old documentation of this UI element: <old_doc>
\end{lstlisting}

\begin{lstlisting}[caption={Task Template}]
You are an agent that is trained to perform some basic tasks on a smartphone. You will be given a 
smartphone screenshot. The interactive clickable UI elements on the screenshot are labeled with tags starting from "c1". 
The interactive scrollable UI elements on the screenshot are labeled with tags starting from "s1".The tag of each 
interactive element is located in the center of the element. Every screenshot I've given you is a screenshot after 
executing the correct action. 

You can call the following functions to control the smartphone:

1. click(element: str)
This function is used to click an UI element shown on the smartphone screen.
"element" is a tag assigned to an UI element shown on the smartphone screen.
A simple use case can be click(c5), which taps the UI element labeled with "c5".

2. input(text_input: str)
This function is used to insert text input in an input field/box. text_input is the string you want to insert and must 
be wrapped with double quotation marks. A simple use case can be text("Hello, world!"), which inserts the string 
"Hello, world!" into the input area on the smartphone screen. This function is usually callable when you see a screenshot 
about text inputing.

3. scroll(element: str, direction: str)
This function is used to scroll an UI element shown on the smartphone screen, usually a scroll view or a slide bar.
"element" is a tag assigned to an UI element shown on the smartphone screen. "direction" is a string that 
represents one of the four directions: up, down, left, right. "direction" must be wrapped with double quotation 
marks.
A simple use case can be swipe(s21, "up"), which scroll up the UI element labeled with "s21".

<ui_document>
The task you need to complete is to <task_description>, to complete this task you should perform current task 
<current_task_desc>. Your past actions to proceed with this task are summarized as follows: <last_act>
Now, given the documentation and the following labeled screenshot, you need to think and call the function needed to 
proceed with the task. Your output should include three parts in the given format:
Observation: <Describe what you observe in the image>
Thought: <To complete the given task, what is the next step I should do>
Action: <The function call with the correct parameters to proceed with the task.>
Summary: <Summarize your past actions along with your latest action in one or two sentences. Do not include the 
tag in your summary>
You can only take one action at a time, so please directly call the function.
\end{lstlisting}
\subsection{Sampling representativeness} \label{sec:sampling}

\begin{table}[t]
\centering
\caption{App counts and proportions in full and subset datasets.}
\scriptsize
\begin{tabular}{lcccc}
\toprule
\textbf{App} & \textbf{Full} & \textbf{Subset} & \textbf{Full (\%)} & \textbf{Sampled (\%)} \\
\midrule
BaiduMap & 52 & 2 & 0.43 & 0.56 \\
QQBrowser & 447 & 13 & 3.72 & 3.67 \\
anjuke & 2773 & 82 & 23.11 & 23.16 \\
baiduBrowser & 682 & 20 & 5.68 & 5.65 \\
bili & 2658 & 78 & 22.15 & 22.03 \\
kugou & 943 & 28 & 7.86 & 7.91 \\
medicinehelper & 831 & 25 & 6.93 & 7.06 \\
taptap & 1570 & 46 & 13.08 & 12.99 \\
vipshop & 893 & 26 & 7.44 & 7.34 \\
ximalaya & 1151 & 34 & 9.59 & 9.60 \\
\bottomrule
\end{tabular}
\label{tab:app_counts}
\vspace{-0.3cm}
\end{table}

To construct a representative evaluation subset, we perform stratified sampling from the full dataset, with application-level proportions preserved. As detailed in Table~\ref{tab:app_counts}, the sampled subset closely mirrors the original distribution across a diverse range of applications. Despite a substantial reduction in data volume, the relative proportions of key apps (e.g., anjuke, bili, taptap) remain nearly identical between the full dataset and the sampled subset (e.g., 23.11\% vs. 23.16\%, 22.15\% vs. 22.03\%). This strong alignment demonstrates the effectiveness of our sampling strategy in maintaining distributional fidelity. By preserving both frequent and less frequent app categories, the subset ensures that evaluation results remain reflective of real-world deployment conditions. Consequently, this sampled dataset serves as a compact yet reliable benchmark for downstream agent performance analysis.

\subsection{Test Case Study}\label{appendix:casestudy}

\textbf{1. Common-Simple}  
\begin{figure*}[!t]
  \centering
  \includegraphics[width=0.95\textwidth]{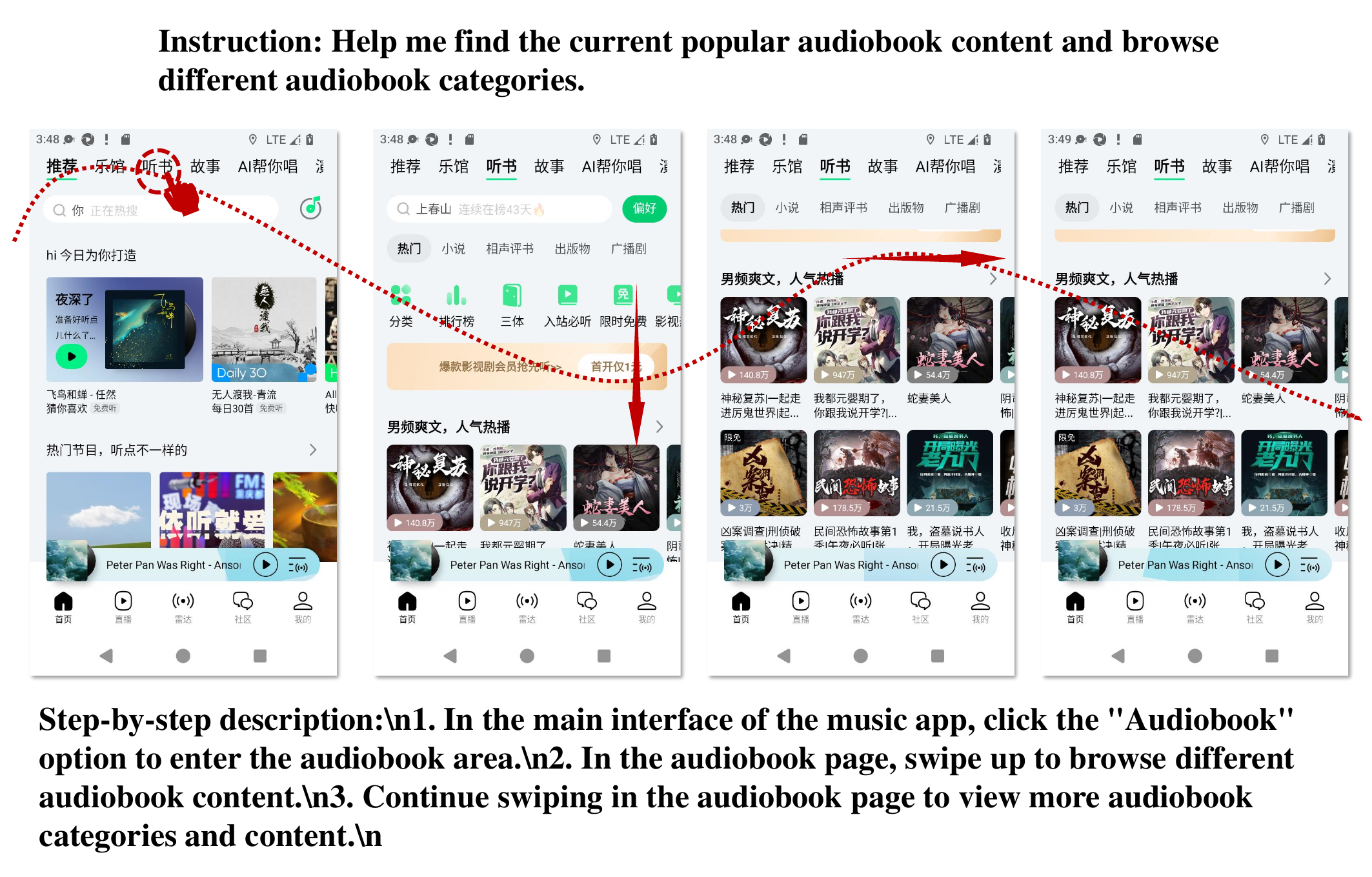}
  \caption{Common-Simple test case. }
  \label{simple_task}
\end{figure*}

\textbf{2. Common-Complex}  

\begin{figure*}[!t]
  \centering
  \includegraphics[width=0.8\textwidth]{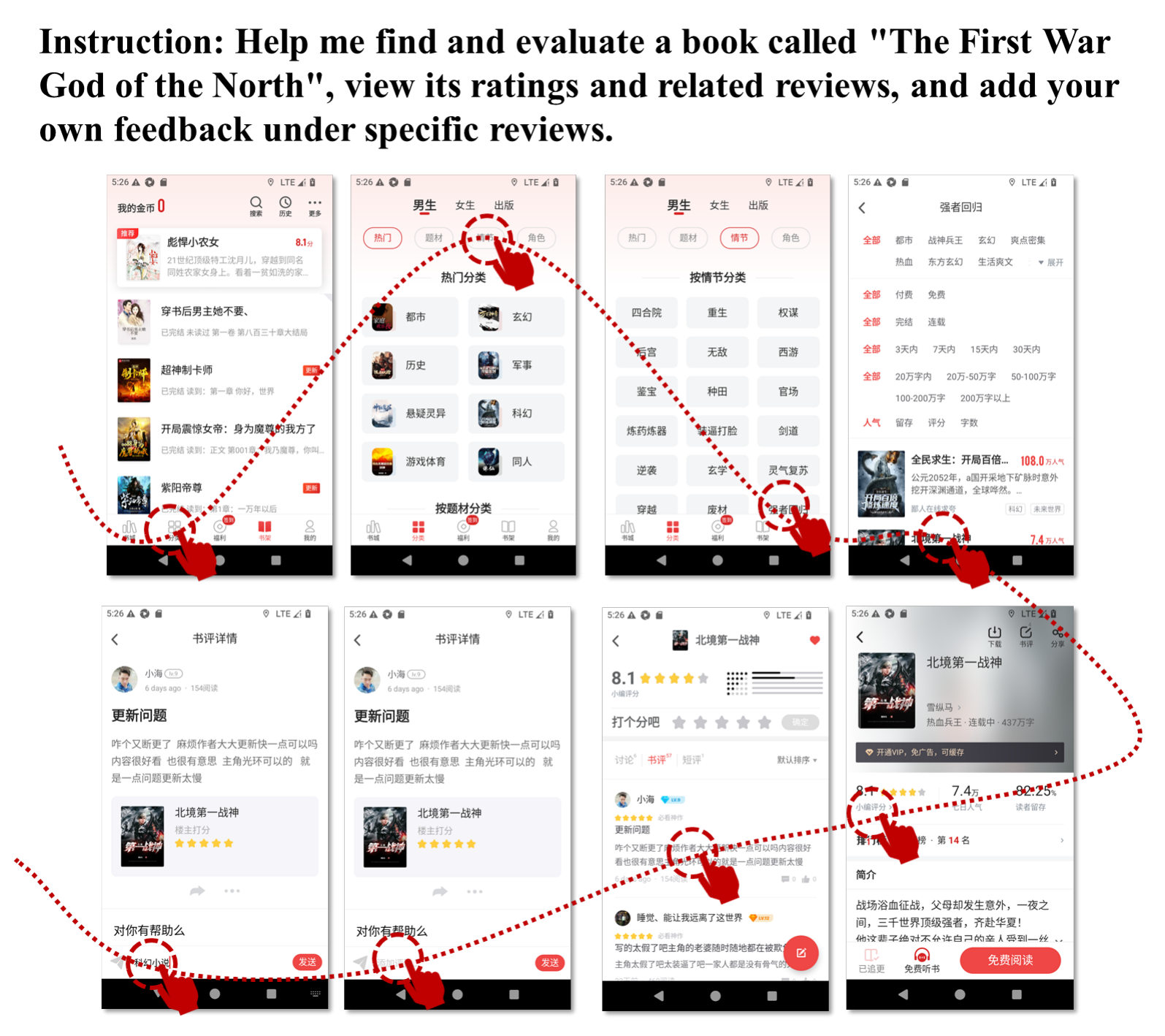}
  \caption{Common-Complex test case. }
  \label{complex_task}
\end{figure*}

Figure \ref{complex_task} shows a Common-Complex case of Mobile-Bench-v2 and the GIAS results are as follows:
\begin{enumerate}[label=\arabic*.]
    \item On the ``My Gold'' page of the mobile app, click the ``Category'' tab to enter the book category page.
    \item On the category page, select the ``Plot'' category under the ``Boys'' tab.
    \item In the plot category, select the ``Return of the Strong'' category to enter the list of books in this category.
    \item In the ``Return of the Strong'' category, select the book ``The First War God of the North.''
    \item On the book details page of ``The First War God of the North,'' click the rating of 8.1 to view the ratings and reviews.
    \item On the review page, click ``Must-see masterpiece'' to view specific book review details.
    \item Enter the comment ``Science Fiction'' on the book review details page and submit it.
\end{enumerate}

Task: Help me find and evaluate a book called \textit{``The First War God of the North''}, view its ratings and related reviews, and add your own feedback under specific reviews.

\textbf{3. Noisy Data}
\begin{figure*}[!t]
  \centering
  \includegraphics[width=0.8\textwidth]{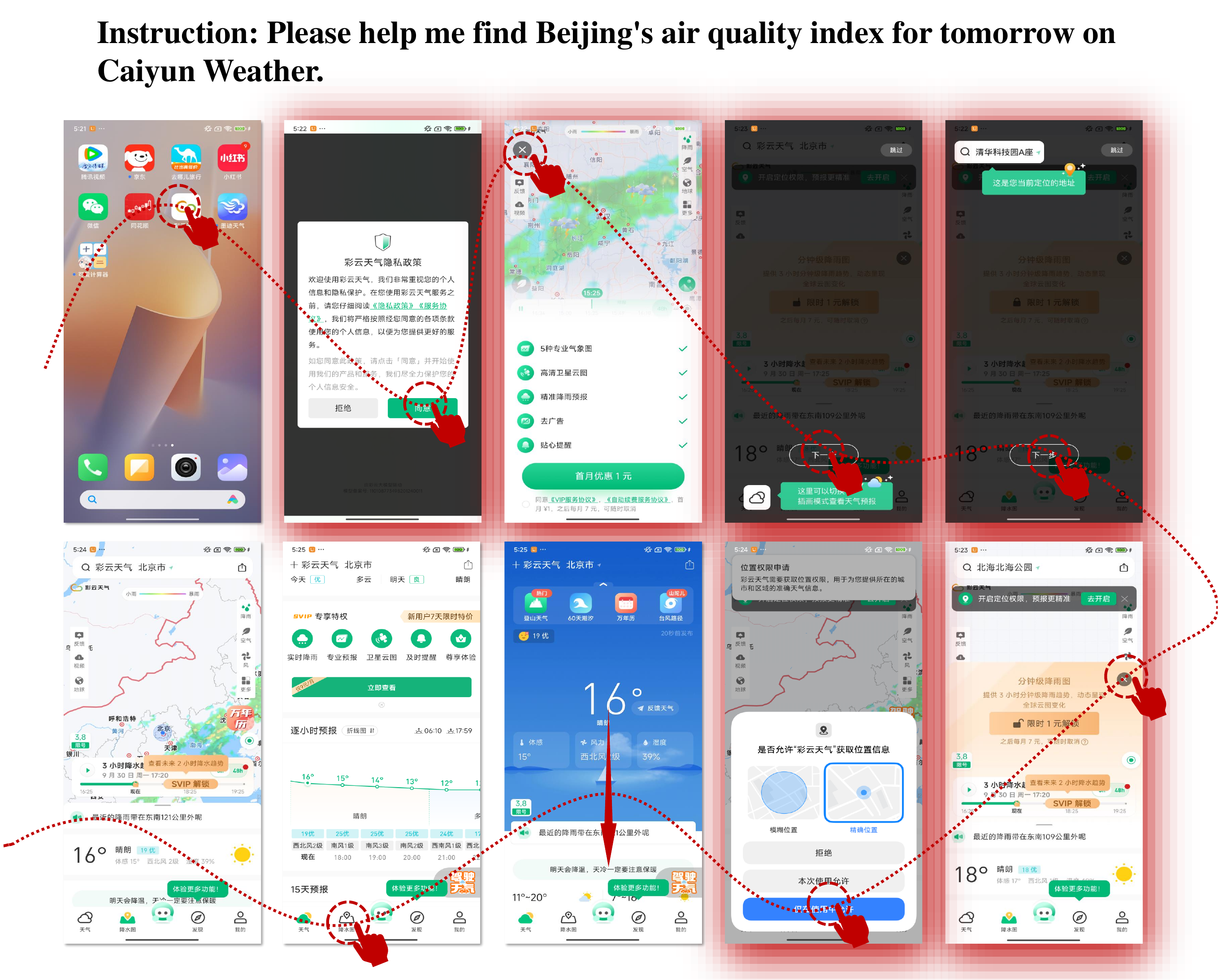}
  \caption{Noisy Data test case. The \textcolor{red}{red shadow} in the GUI screenshot are advertisements, pop-ups, or tutorial noise steps.}
  \label{noisy_mobile}
\end{figure*}

\textbf{4. AITZ-Noisy}
\begin{figure*}[!t]
  \centering
  \includegraphics[width=1\textwidth]{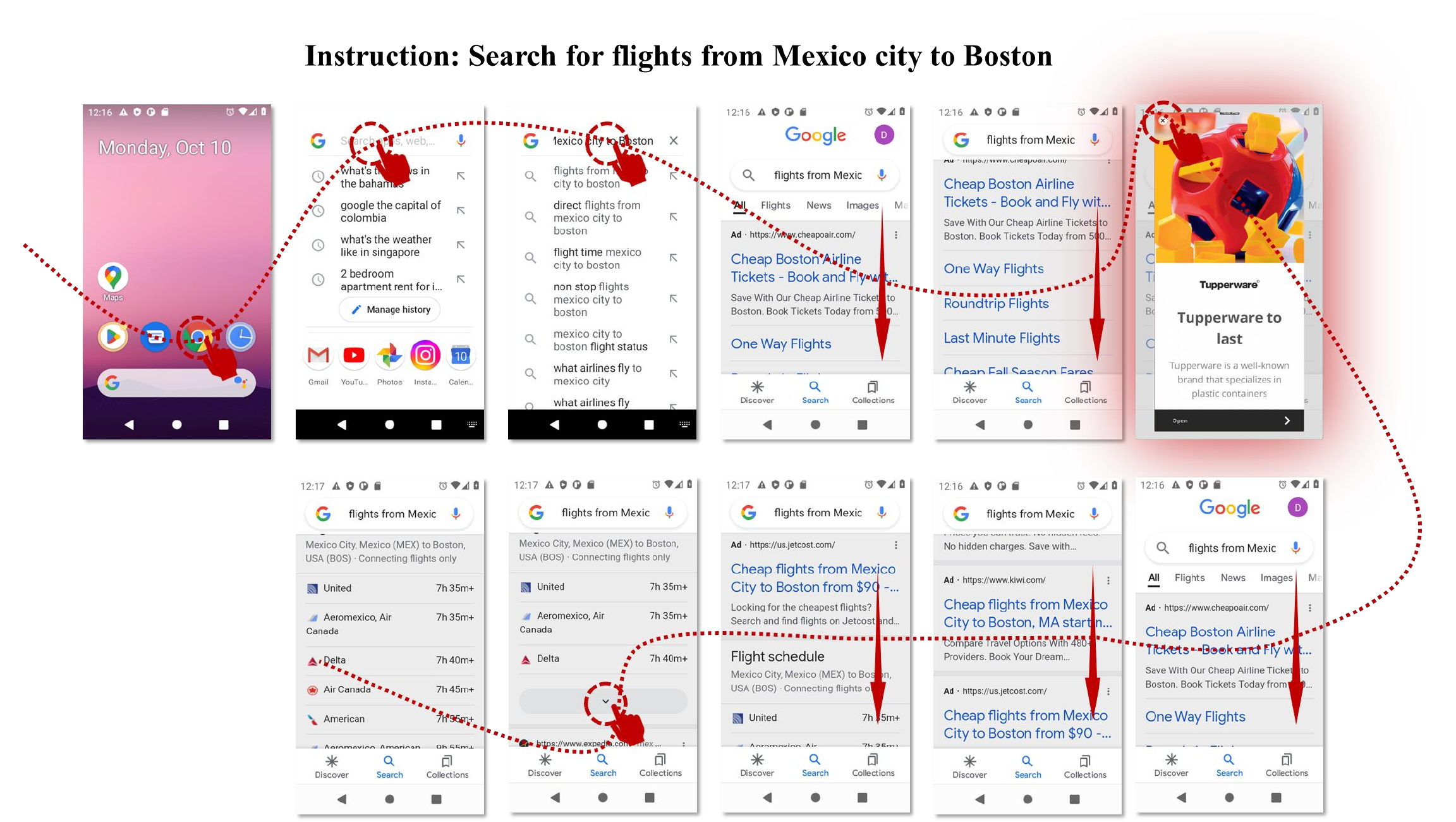}
  \caption{AITZ-Noisy test case. The \textcolor{red}{red shadow} GUI screenshot in the trajectory is artificially inserted noise. }
  \label{aitz_noise}
\end{figure*}

\textbf{5. Ambiguous Data}\label{appendix:Ambiguous Data}
\begin{table}[t!]
\centering
\caption{Test cases study on Mobile-Bench-Ambiguous. }
\resizebox{0.9\linewidth}{!}{
\begin{tabular}{p{6cm}p{8cm}@{}}
\toprule
\cellcolor{gray!20} \textbf{Ambiguous Instruction} & \cellcolor{gray!20} Find information about a movie \\
\midrule
Q: Which app should be used?  & A: Use Douban. \\ 
Q: Which app does this page belong to? & A: Douban.   \\ 
Q: In which section should I search?         & A: Movies.  \\ 
Q: Do you want to browse a specific ranking? & A: Yes.     \\ 
Q: For which time period?   & A: Upcoming releases.        \\ 
Q: How should it be ranked? & A: By popularity.            \\ 
Q: Which movie do you want to check?& A: The most popular upcoming movie.  \\ Q: What information do you need?& A: A complete summary. \\ 
 \midrule
 \cellcolor{gray!20} \textbf{Full Instruction} & \cellcolor{gray!20} Find the most popular upcoming movie on Douban   \\
 \midrule
 \cellcolor{gray!20} \textbf{Ambiguous Instruction} & \cellcolor{gray!20} Find a midnight snack \\
 \midrule
Q: Which app should be used?  & A: Use Ele.me.            \\
Q: Would you like to filter by specific snack categories, speed,        & A: Find new items from the nearest store that can    \\
Q: Any other conditions?       & A: deliver within 30 minutes.          \\ 
Q: Do you have a specific price range?     & A: No specific price range.            \\ 
Q: Do you have a preferred cuisine or taste?             & A: No preference, just quick delivery within 30 minutes.           \\ 
Q: Which section should you search?         & A: Food delivery.        \\ 
Q: What are the speed requirements?         & A: Within 30 minutes.    \\ 
Q: What are the distance requirements?      & A: Nearest store.        \\ 
\midrule
\cellcolor{gray!20} \textbf{Full Instruction} & \cellcolor{gray!20} Find a new delivery item from the nearest store on Ele.me that can deliver within 30 minutes.  \\
\bottomrule
\end{tabular}
}
\label{tab:presetting_qa}
\end{table}
\section{Limitations} \label{limitation}
Although multi-path validation similar to that on online machines was achieved on Mobile-Bench-v2, the diverse range of text inputs cannot be exhaustively covered, which differentiates it from online machines. Advanced agents such as AutoGLM \citep{liu2024autoglm} and others deployed by smartphone manufacturers could not be tested due to permission restrictions.

\section{Ethics Statement} \label{ethics}
We have rigorously refined our dataset to remove any elements that could compromise personal privacy, thereby guaranteeing the highest level of protection for individual data. All data annotations were completed by crowdsourced volunteers, to whom we paid \$0.5 per step as compensation and provided the necessary training. The human evaluation of our work was carried out through a meticulously randomized selection of IT professionals. This process ensured a gender-balanced and educationally diverse panel, reflecting a wide spectrum of perspectives and expertise.

\newpage

\end{document}